\documentclass[12pt,letterpaper]{article}

\usepackage[T1]{fontenc}
\usepackage[utf8]{inputenc}
\usepackage{textcomp,lmodern,microtype}
\RequirePackage{amsmath,amsthm,amsfonts,amssymb}
\RequirePackage{graphicx,booktabs,array,multirow,colortbl,xcolor}
\RequirePackage{enumitem}
\setlist{topsep=4pt plus 2pt minus 1pt, itemsep=2pt plus 1pt,
  parsep=0pt, partopsep=0pt}
\RequirePackage[authoryear]{natbib}
\usepackage{setspace}
\usepackage{etoolbox}
\makeatletter
\patchcmd{\section}{-3.5ex \@plus -1ex \@minus -.2ex}{-2.5ex \@plus -.5ex \@minus -.2ex}{}{}
\patchcmd{\section}{2.3ex \@plus .2ex}{1.5ex \@plus .2ex}{}{}
\patchcmd{\subsection}{-3.25ex\@plus -1ex \@minus -.2ex}{-2ex \@plus -.5ex \@minus -.2ex}{}{}
\patchcmd{\subsection}{1.5ex \@plus .2ex}{1ex \@plus .2ex}{}{}
\makeatother

\appto\normalsize{%
  \setlength{\abovedisplayskip}{8pt plus 2pt minus 2pt}%
  \setlength{\belowdisplayskip}{8pt plus 2pt minus 2pt}%
  \setlength{\abovedisplayshortskip}{3pt plus 2pt}%
  \setlength{\belowdisplayshortskip}{6pt plus 2pt minus 2pt}%
}
\RequirePackage{xr-hyper}
\RequirePackage[colorlinks,linkcolor=blue,citecolor=blue,urlcolor=blue]{hyperref}
\usepackage{xurl}

\numberwithin{equation}{section}
\theoremstyle{plain}
\newtheorem{theorem}{Theorem}
\newtheorem{corollary}{Corollary}[section]
\newtheoremstyle{boldremark}
  {.5\topsep}{.5\topsep}
  {\normalfont}{}
  {\bfseries}{.}{5pt plus 1pt minus 1pt}
  {\thmname{#1}\thmnumber{ #2}\thmnote{ {\normalfont(#3)}}}
\theoremstyle{boldremark}
\newtheorem{remark}{Remark}
\theoremstyle{definition}
\newtheorem{definition}{Definition}
\newtheorem{assumption}{Assumption}
\theoremstyle{plain}
\newtheorem{proposition}[theorem]{Proposition}

\providecommand{\tblval}[2]{#1\,\scalebox{0.7}{(#2)}}

\usepackage[section]{placeins}

\begin{document}
\def\spacingset#1{\renewcommand{\baselinestretch}{#1}\small\normalsize}
\spacingset{1}
\title{\bf Prediction-Powered Regression via Calibration: Beyond the Supervised Minimax Benchmark}
\author{Jin Su\\
  {\footnotesize\href{mailto:jin2930.su@polyu.edu.hk}{\texttt{jin2930.su@polyu.edu.hk}}}\\
  \parbox[t]{0.27\textwidth}{\centering\footnotesize Department of Applied Mathematics, The Hong Kong Polytechnic University}
  \and
  Yong Zhou\\
  {\footnotesize\href{mailto:yzhou@fem.ecnu.edu.cn}{\texttt{yzhou@fem.ecnu.edu.cn}}}\\
  \parbox[t]{0.27\textwidth}{\centering\footnotesize School of Statistics, East China Normal University}
  \and
  Jian Huang\\
  {\footnotesize\href{mailto:j.huang@polyu.edu.hk}{\texttt{j.huang@polyu.edu.hk}}}\\
  \parbox[t]{0.27\textwidth}{\centering\footnotesize Department of Applied Mathematics and Department of Data Science and Artificial Intelligence, The Hong Kong Polytechnic University}}
\date{}
\maketitle
\bigskip
\begin{abstract}
We study nonparametric regression with limited labeled data and abundant observations carrying potentially inaccurate pseudo-labels.
Our proposed method, weighted prediction-powered regression via calibration (wPPRC), first calibrates pseudo-labels using labeled data, then fits a predictor by weighted regression using observed and calibrated responses.
Over H\"older classes, we derive nonasymptotic excess-risk bounds for clipped ReLU estimators and establish a minimax lower bound for the corresponding regression problem. 
Under suitable conditions, the upper and lower bounds match up to logarithmic factors. 
Moreover, wPPRC can achieve a worst-case risk of smaller order than the minimax risks under four baseline observation structures on the same joint distribution class, demonstrating the value of additional pseudo-labels.
We also construct a validation-selected (VS) estimator using an independent labeled validation sample to choose the weight. 
The VS estimator satisfies an expected-risk oracle inequality and preserves the improvement over the baseline minimax risks with the same total labeled-data budget. Simulations and an application to predicting homeless counts illustrate the gains of the proposed estimators.
\end{abstract}

\par\medskip
\spacingset{1.8}
\clearpage
\section{Introduction}

In many applications, covariates are available at scale, but high-quality outcomes are scarce.
Obtaining such outcomes may require costly experiments \citep{xue2016accelerated}, long-term
follow-up \citep{aiyegbusi2022key}, or expert annotation \citep{radsch2023labelling}.
By contrast, machine-learning systems can generate
pseudo-labels for large datasets at relatively low cost. Examples include text annotations produced by large language models \citep{gilardi2023chatgpt} and medical-image labels generated by domain-specific AI systems with limited expert review \citep{kim2022accurate}.
Recent advances in tabular foundation models also make predictions available for tabular regression tasks with limited labeled data \citep{hollmann2025accurate,qu2026tabiclv2}.
These pseudo-labels offer an abundant but potentially inaccurate source of response information.

In this paper, we study estimation of the regression function
\(f^*(x)=\mathbb{E}(Y\mid X=x)\), where
\(X\in\mathcal X\subset\mathbb R^{d_x}\) is the covariate and
\(Y\in\mathcal Y\subset\mathbb R\) is the true response. Let
\(\widetilde Y\) denote a model-generated pseudo-label. Consider two independent samples:
\[
\mathcal L
=
\{(X_i,Y_i,\widetilde Y_i)\}_{i=1}^{N_{\ell}}
\overset{\mathrm{iid}}{\sim}P_{X,Y,\widetilde Y},
\qquad
\mathcal U
=
\{(X_j,\widetilde Y_j)\}_{j=1}^{N_u}
\overset{\mathrm{iid}}{\sim}P_{X,\widetilde Y}.
\]
Here, \(\mathcal L\) is the labeled sample and \(\mathcal U\) the pseudo-labeled sample.
We focus on the setting where \(N_\ell/N_u\to0\) as \(N_\ell,N_u\to\infty\).
The central question is when additional pseudo-labels can  improve the attainable accuracy of regression estimation.

Two broad lines of work provide complementary perspectives on this problem.
The first, prediction-powered inference (PPI), uses a small sample of gold-standard outcomes to correct estimates based on abundant model predictions, enabling valid and potentially more efficient inference \citep{angelopoulos2023prediction}.
PPI++ adapts the weight assigned to model predictions to their quality \citep{angelopoulos2023ppi++}; recalibrated PPI uses an imputed loss \citep{ji2025predictions}; and conditional extensions address functionals at fixed covariate values \citep{sui2026prediction}. Our focus is estimation of the entire regression function and the conditions under which auxiliary predictions improve its accuracy.

The second line, semi-supervised learning (SSL), combines a small labeled sample with a larger \(X\)-only sample \citep{liang2007use}. This approach can improve estimation of means \citep{zhang2019semi,zhang2022highdimensional}, linear projections \citep{chakrabortty2018efficient,azriel2022semi}, and general \(M\)-estimation targets \citep{song2024general}. Recent work also develops optimal and safe procedures for high-dimensional problems \citep{deng2024optimal}. For estimation of the entire regression function, gains from covariates alone generally require structure linking their marginal distribution to the target \citep{wasserman2007statistical,azizyan2013density}, as in manifold learning \citep{moscovich2017minimax} or population Sobolev penalties \citep{ding2025semi}.
In our setting, the additional sample also includes pseudo-labels, and we study the value of the response information they carry beyond \(X\).

We characterize this information through the augmented conditional mean
\[
\mu^*(x,\tilde y)
=
\mathbb{E}(Y\mid X=x,\widetilde Y=\tilde y).
\]
The function \(\mu^*\) describes how the pseudo-label informs the conditional mean of the response beyond the covariates. Learning this relationship allows calibration to use that information even when the pseudo-label is biased. The tower property links \(\mu^*\) to the target:
\begin{equation}
f^*(x)=\mathbb{E}\{\mu^*(X,\widetilde Y)\mid X=x\}.
\label{eq:transfer-identity}
\end{equation}
Two extreme cases illustrate the role of pseudo-label information. If \(\widetilde Y=Y\), then \(\mu^*(x,\tilde y)=\tilde y\), and the pseudo-label contains full response information. If \(Y\) and \(\widetilde Y\) are conditionally independent given \(X\), then \(\mu^*(x,\tilde y)=f^*(x)\), and the pseudo-label carries no response information beyond the covariates.

Several existing methods use auxiliary predictions or information to improve estimation.
\citet{mccaw2024synthetic} develop a synthetic surrogate method that uses
predicted outcomes to improve estimation of finite-dimensional regression
coefficients.
\citet{xia2024prediction} propose PAST, which predicts
unlabeled responses from covariates and auxiliary information, then combines these predictions with labeled data to learn a regression function of the covariates alone.
\citet{shi2026coupled} propose coupled training, which jointly fits a deployment predictor and an auxiliary predictor using privileged training information.
Section~\ref{sec:connections-existing-methods} discusses those methods in more detail.
Related work also considers settings where black-box predictions remain available at deployment. \citet{li2026personalizing} estimate a bias correction through local smoothing, while \citet{zhou2026blackbox} models the residuals between observed responses and black-box predictions and uses validation to select between the original and corrected predictors.  Taken together, these works illustrate both the potential and the challenges of using auxiliary predictions in regression problems.

Building on \eqref{eq:transfer-identity} and the calibration principle underlying PAST, we consider weighted prediction-powered regression via calibration (wPPRC), a family of two-step estimators.
The first step calibrates pseudo-labels by estimating \(\mu^*\) from the labeled sample.
The second step then regresses observed responses and calibrated pseudo-labels on \(X\), with weight \(\lambda\in[0,1]\) controlling their relative contributions.
Calibration can improve estimation when the relationship between \(Y\) and \((X,\widetilde Y)\) is easier to estimate than that between \(Y\) and \(X\) alone.
We also develop a validation-selected (VS) estimator that uses an independent labeled validation sample to select among candidates fitted with different weights.

Our theoretical analysis addresses three main questions:
\begin{enumerate}[label=(\roman*), leftmargin=1.6em, labelsep=0.25em]
		\item \textit{When is wPPRC minimax optimal?}
	Under H\"older smoothness and conditionally sub-Gaussian response noise assumptions, we derive risk bounds for wPPRC with clipped ReLU estimators that account for approximation and statistical errors
	(Theorem~\ref{thm:weighted-holder-risk} and Corollary~\ref{cor:weighted-convergence-rate})
	and establish a minimax lower bound for estimating the target regression function
	(Theorem~\ref{thm:minimax-lower-main}).
	We identify when wPPRC attains the minimax rate up to logarithmic factors and when supervised estimation suffices
	(Section~\ref{sec:minimax-optimality}).

	\item \textit{When do additional pseudo-labels improve estimation?}
We compare the risk upper bound for wPPRC with the minimax risks under four baseline observation structures on the same joint distribution class, differing in access to pseudo-labels on the labeled sample and to additional covariates (Table~\ref{tab:reference-lower-bounds}).
When the calibration function is sufficiently smooth relative to the target regression function and the pseudo-labeled sample grows polynomially faster than the labeled sample, the risk upper bound for wPPRC is of strictly smaller order than the minimax risks of all four baselines (Theorem~\ref{thm:baseline-minimax-separation}).
This improvement holds even against a baseline with the same labeled triples and all additional covariates, highlighting the value of pseudo-labels on the additional observations.

\item \textit{Can validation selection retain these gains while controlling risk relative to supervised estimation?}
We establish an expected-risk oracle inequality for the VS estimator, comparing its risk with that of the best candidate and quantifying the cost of validation selection
(Proposition~\ref{prop:validation-selection-optimized}).
When the number of candidates remains fixed and the validation and labeled training sample sizes are of the same order, including supervised regression among the candidates allows validation selection to preserve the convergence rate established for supervised regression.
The VS estimator also retains the improvement over all four baseline minimax risks, even when the baselines receive the same total labeled-data budget
(Corollary~\ref{cor:validation-minimax-separation}).
\end{enumerate} 
In addition to the theoretical analysis, we assess the finite-sample performance of wPPRC and VS estimators through simulations and a homeless-count prediction study using pseudo-labels from tabular foundation models.

The remainder of the paper is organized as follows.
Section~\ref{sec:prediction-powered-estimator} develops the wPPRC and VS estimators.
Section~\ref{sec:convergence-analysis} derives risk bounds for these estimators.
Section~\ref{sec:minimax-bounds} establishes minimax lower bounds and compares risks across observation structures.
Section~\ref{sec:linear-regression-weighting} examines a linear special case and derives the variance-optimal weight.
Section~\ref{sec:connections-existing-methods} compares wPPRC with existing methods.
Sections~\ref{sec:simulation-studies} and~\ref{sec:real-data-analysis} present simulation and real-data studies, respectively.
Section~\ref{sec:conclusion} concludes.
Technical proofs, additional numerical results, and implementation details are provided in the Supplementary Material (SM).

\smallskip
\noindent\textbf{Notation.} Constants \(c,C>0\) are independent of sample sizes and may vary between lines. For nonnegative sequences, \(a_n\lesssim b_n\) means \(a_n\le Cb_n\), \(a_n\gtrsim b_n\) means \(b_n\lesssim a_n\), and \(a_n\asymp b_n\) means both relations hold. We use the standard \(O\) and \(o\) notation, and write \(O_P\) and \(o_P\) for their stochastic counterparts. Let \(a\vee b=\max\{a,b\}\).
We write \(\mathbb N_0=\{0,1,2,\ldots\}\), and let \(\lfloor t\rfloor\) and \(\lceil t\rceil\) denote the floor and ceiling of \(t\), respectively. The symbol \(I_d\) denotes the \(d\times d\) identity matrix, \(\perp\) denotes independence, and \(\log\) denotes the natural logarithm.
For a finite set \(A\), \(|A|\) denotes its cardinality. The notation \(\mathbf1\{E\}\) denotes the indicator of an event \(E\), while \(\mathbf1\) denotes an all-ones vector.
The superscript \(\top\) denotes transpose. For vectors, \(\|\cdot\|_2\) is the Euclidean norm; for vectors and matrices, \(\|\cdot\|_\infty\) is the largest absolute entry and \(\|\cdot\|_0\) counts nonzero entries. For a function \(g\), \(\|g\|_\infty\) denotes its supremum norm, and \(\|g\|_{L^2(P)}^2=\int |g|^2\,dP\) for a probability measure \(P\).
\section{Prediction-Powered Estimation via Calibration} 
\label{sec:prediction-powered-estimator}

We introduce the wPPRC family by first calibrating pseudo-labels using labeled data and then fitting a predictor through weighted regression on observed and calibrated responses. We specify ReLU network classes for both steps and describe how to select the weight using an independent labeled validation sample.

\subsection{Two-Step Construction}

Write \(N=N_\ell+N_u\) and fix a deterministic weight \(\lambda\in[0,1]\), possibly depending on the sample sizes. Based on \eqref{eq:transfer-identity}, we construct a two-step estimator using two working function classes:
\(\mathcal H_{\mathrm{in}}\) for calibration functions of \((x,\tilde y)\) and \(\mathcal H_{\mathrm{out},\lambda}\)
for prediction functions of \(x\).
The subscripts \(\mathrm{in}\) and \(\mathrm{out}\) refer to the first-
and second-step regressions, with input dimensions
\(d_{\mathrm{in}}=d_x+1\) and \(d_{\mathrm{out}}=d_x\), respectively.
The output class \(\mathcal H_{\mathrm{out},\lambda}\) may depend on \(\lambda\). For a fixed \(\lambda\), we write \(\mathcal H_{\mathrm{out}}\) when no confusion arises.
\begin{itemize}[leftmargin=1.5em, labelsep=0.5em, topsep=2pt, before={\predisplaypenalty=0}]
\item  \textbf{Step 1: Calibration with \((X,\widetilde Y)\).} Using the labeled sample \(\mathcal L\), estimate the augmented conditional mean \(\mu^*\) by least squares:
\begin{align*}
\widehat\mu
\in
\arg\min_{\mu\in\mathcal{H}_{\mathrm{in}}}
\frac1{N_{\ell}}\sum_{i=1}^{N_{\ell}}
\{Y_i-\mu(X_i,\widetilde{Y}_i)\}^2 .
\end{align*}

\item \textbf{Step 2: Weighted regression on \(X\).}
Evaluate \(\widehat\mu\) on \(\mathcal U\) and combine the calibrated predictions with the labeled responses:
\begin{align*}
\widehat Q_\lambda(f)
&=(1-\lambda)\frac1{N_u}\sum_{j=1}^{N_u}
\{\widehat\mu(X_j,\widetilde Y_j)-f(X_j)\}^2
+\lambda\frac1{N_\ell}\sum_{i=1}^{N_\ell}\{Y_i-f(X_i)\}^2,
\\
\widehat f_\lambda&\in\arg\min_{f\in\mathcal H_{\mathrm{out},\lambda}}\widehat Q_\lambda(f).
\end{align*}
\end{itemize}
The procedure defines the weighted prediction-powered regression via calibration (wPPRC) family \(\{\widehat f_\lambda:\lambda\in[0,1]\}\).
Calibration accommodates nonlinear dependence on both inputs without requiring unbiased pseudo-labels.
For fixed calibration \(\mu\), the second-step population objective seeks the best approximation within \(\mathcal H_{\mathrm{out},\lambda}\) to
\(g_{\lambda,\mu}:=\lambda f^*+(1-\lambda)\mathcal T\mu\), where \((\mathcal T\mu)(x)=\mathbb E\{\mu(X,\widetilde Y)\mid X=x\}\). When \(\mu=\mu^*\), this target equals \(f^*\).

The choice of \(\lambda\) determines how the labeled responses and calibrated unlabeled responses enter the second step.
For example, at \(\lambda=0\), the second step regresses the calibrated responses on \(X\) using only the unlabeled sample. At \(\lambda=N_\ell/N\), the objective becomes pooled least squares on the observed labeled responses and the calibrated unlabeled responses, with equal weight assigned to each observation. At \(\lambda=1\), the estimator reduces to supervised least squares, denoted by \(\widehat f_{\mathrm{sup}}:=\widehat f_1\), and the calibration step is omitted.

We use the following class of ReLU networks for both \(\mathcal H_{\mathrm{in}}\) and \(\mathcal H_{\mathrm{out}}\).

\begin{definition}[ReLU network class]
\label{def:relu-network}
	Let $\operatorname{ReLU}(t)=\max\{t,0\}$, applied
	componentwise to vectors. For $U\in(0,\infty)$, define
	$\operatorname{clip}_U(t)=\min\{\max(t,-U),U\}$, and set
	$\operatorname{clip}_\infty(t)=t$.
	For integers $d,W,S\geq1$ and $L\geq2$, and for $R>0$ and
	$U\in(0,\infty]$, let $\mathcal F_{d,L,W,S,R,U}$ be the class of
	functions $f_\phi:\mathbb R^d\to\mathbb R$ of the form
	$f_\phi(x)=\operatorname{clip}_U\{\tilde f_\phi(x)\}$, where
	$\tilde f_\phi=L_L\circ\operatorname{ReLU}\circ L_{L-1}\circ\operatorname{ReLU}
	\circ\cdots\circ\operatorname{ReLU}\circ L_1$,
	$L_\ell(z)=A_\ell z+b_\ell$, and
	$\phi=\{(A_\ell,b_\ell)\}_{\ell=1}^L$.
	The layer parameters satisfy
	$A_1\in\mathbb R^{W\times d}$, $b_1\in\mathbb R^W$,
	$A_\ell\in\mathbb R^{W\times W}$, $b_\ell\in\mathbb R^W$ for
	$1<\ell<L$, and $A_L\in\mathbb R^{1\times W}$, $b_L\in\mathbb R$.
	The network parameters satisfy
	$\sum_{\ell=1}^{L}\{\|A_\ell\|_0+\|b_\ell\|_0\}\leq S$ and
	$\max_{1\leq\ell\leq L}
	\{\|A_\ell\|_\infty,\|b_\ell\|_\infty\}\leq R$.
\end{definition}
For the two estimation steps, we set
\begin{align}
	\label{eq:nn-working-classes}
	\mathcal{H}_{\mathrm{in}}
	=
	\mathcal F_{d_{\mathrm{in}},L_{\mathrm{in}},W_{\mathrm{in}},S_{\mathrm{in}},R_{\mathrm{in}},U_{\mathrm{in}}},
	\qquad
	\mathcal{H}_{\mathrm{out}}
	=
	\mathcal F_{d_{\mathrm{out}},L_{\mathrm{out}},W_{\mathrm{out}},S_{\mathrm{out}},R_{\mathrm{out}},U_{\mathrm{out}}}.
\end{align}
In particular, \(\mathcal H_{\mathrm{out},1}\) is the working class for supervised estimation.

\subsection{Choosing the Weight by Validation}
\label{sec:weight-selection}
The weight \(\lambda\) may be specified in advance or selected using an independent labeled validation sample \(\mathcal L_{\mathrm{val}}\) of size \(N_{\ell,\mathrm{val}}\) from the same population.
(This sample may be obtained by randomly splitting the available labeled data; in that case, \(\mathcal L\) denotes the fitting subset and \(N_\ell\) its size.)
For a prespecified nonempty finite set \(\Lambda\subset[0,1]\), construct \(\widehat f_\lambda\) for each \(\lambda\in\Lambda\) using the procedure above. Select
\begin{equation*}
\widehat{\lambda}\in\arg\min_{\lambda\in\Lambda}
\frac1{N_{\ell,\mathrm{val}}}\sum_{(X_i,Y_i,\widetilde Y_i)\in\mathcal L_{\mathrm{val}}}
\{Y_i-\widehat f_{\lambda}(X_i)\}^2,
\end{equation*}
and return the fitted candidate \(\widehat f_{\widehat\lambda}\). We refer to \(\widehat f_{\widehat\lambda}\) as the validation-selected (VS) estimator. In practice, we do not know in advance whether pseudo-labels will improve prediction, so we can include \(1\) in \(\Lambda\) to allow supervised regression as a candidate.

\section{Risk Bounds for wPPRC and VS Estimators}
\label{sec:convergence-analysis}
In this section, we derive risk bounds for estimators with fixed or validation-selected weights. We first introduce the H\"older class \citep{tsybakov2009} used to describe the smoothness of the target functions.

\begin{definition}[H\"older class]
\label{def:holder}
Let $\Omega\subset\mathbb R^d$ be a fixed compact rectangle with positive side lengths. For \({\beta}>0\), set $m=\lceil {\beta}\rceil-1$.
For a multi-index $\nu\in\mathbb N_0^d$, write
$|\nu|=\sum_{j=1}^{d}\nu_j$ and let $D^{\nu}$ denote the corresponding partial derivative, with $D^0g=g$. For functions \(g\) whose derivatives $D^{\nu}g$,
$|\nu|\le m$, exist in the interior of \(\Omega\) and extend continuously to
$\Omega$, define
\[
\|g\|_{C^{\beta}(\Omega)}
=\sum_{|\nu|\le m}\|D^{\nu}g\|_{\infty}
+\sum_{|\nu|=m}
\sup_{\substack{x,x'\in\Omega\\x\ne x'}}
\frac{|D^{\nu}g(x)-D^{\nu}g(x')|}{\|x-x'\|_2^{{\beta}-m}}.
\]
Let $C^{\beta}(\Omega)$ be the class of such functions with finite norm. For $B>0$,
define the H\"older ball
\[
C^{\beta}(\Omega,B)=\{g\in C^{\beta}(\Omega):\|g\|_{C^{\beta}(\Omega)}\le B\}.
\]
\end{definition}

Recall that \(d_{\mathrm{out}}=d_x\) and \(d_{\mathrm{in}}=d_x+1\). Let
\(\Omega_{\mathrm{out}}=\mathcal X\subset\mathbb R^{d_{\mathrm{out}}}\) and
\(\Omega_{\mathrm{in}}=\mathcal X\times\widetilde{\mathcal Y}\subset\mathbb R^{d_{\mathrm{in}}}\)
be fixed compact rectangular domains with positive side lengths.
For $a\in\{\mathrm{in},\mathrm{out}\}$, let
${\beta}_a>0$ and \(B_a>0\) be fixed smoothness orders and H\"older radii, respectively. The following assumptions specify the model class for our theoretical analysis.

\begin{assumption}[H\"older smoothness]
\label{ass:holder-smoothness}
The conditional mean functions $\mu^*$ and $f^*$ satisfy
$\mu^*\in
C^{{\beta}_{\mathrm{in}}}
(\Omega_{\mathrm{in}},B_{\mathrm{in}})$, $f^*\in
C^{{\beta}_{\mathrm{out}}}
(\Omega_{\mathrm{out}},B_{\mathrm{out}})$.
\end{assumption}

\begin{assumption}[Conditionally sub-Gaussian response noise]
\label{ass:subgaussian-noise}
Let \(\varepsilon=Y-\mu^*(X,\widetilde Y)\). There is a fixed \(\sigma>0\) such that, for every \(t\in\mathbb R\),
$\mathbb E(e^{t\varepsilon}\mid X,\widetilde Y)
\le e^{\sigma^2t^2/2}$ almost surely.
\end{assumption}

Assumption~\ref{ass:holder-smoothness} imposes separate smoothness conditions on \(\mu^*\) and \(f^*\), since the regularity of \(f^*\) also depends on how the conditional distribution of \(\widetilde Y\) varies with \(X\).
Assumption~\ref{ass:subgaussian-noise} allows unbounded responses while, together with bounded \(\mu^*\), ensuring \(\mathbb E(Y^2)<\infty\) and finite squared-loss risks for bounded predictors. It also yields conditionally sub-Gaussian noise \(Y-f^*(X)\) for the supervised estimator.

\subsection{Risk Bounds for wPPRC Estimators}
\label{sec:relu-upper-bound}

For a fixed weight \(\lambda\in[0,1]\), define
\[
e_\lambda=\|\widehat f_\lambda-f^*\|_{L^2(P_X)}^2,
\qquad \mathcal R_\lambda=\mathbb E e_\lambda,
\]
where the expectation is over the fitting samples. 

Using the population target defined above, \(g_{\lambda,\widehat\mu}-f^*=(1-\lambda)\mathcal T(\widehat\mu-\mu^*)\). The squared triangle inequality and conditional Jensen's inequality therefore give
\begin{align*}
\mathcal R_\lambda \le 2\mathbb E\|\widehat f_\lambda-g_{\lambda,\widehat\mu}\|_{L^2(P_X)}^2
+2(1-\lambda)^2\mathbb E\|\widehat\mu-\mu^*\|_{L^2(P_{X,\widetilde Y})}^2.
\end{align*}
The two terms represent second-step regression error and scaled calibration error.

We decompose the second-step error as 
\begin{align*}
\mathbb E\|\widehat f_\lambda-g_{\lambda,\widehat\mu}\|_{L^2(P_X)}^2
&=\mathbb E\inf_{f\in\mathcal H_{\mathrm{out},\lambda}}
\|f-g_{\lambda,\widehat\mu}\|_{L^2(P_X)}^2\\
&\quad+\mathbb E\left[
\|\widehat f_\lambda-g_{\lambda,\widehat\mu}\|_{L^2(P_X)}^2
-\inf_{f\in\mathcal H_{\mathrm{out},\lambda}}
\|f-g_{\lambda,\widehat\mu}\|_{L^2(P_X)}^2
\right].
\end{align*}
The approximation term satisfies
\begin{align*}
\mathbb E\inf_{f\in\mathcal H_{\mathrm{out},\lambda}}\|f-g_{\lambda,\widehat\mu}\|_{L^2(P_X)}^2\le 2\inf_{f\in\mathcal H_{\mathrm{out},\lambda}}\|f-f^*\|_{L^2(P_X)}^2
+2(1-\lambda)^2\mathbb E\|\widehat\mu-\mu^*\|_{L^2(P_{X,\widetilde Y})}^2.
\end{align*}
Substituting these expressions into the risk bound and similarly decomposing the calibration error yields
\begin{align*}
\mathcal R_\lambda\le{}&
6(1-\lambda)^2
\underbrace{\inf_{\mu\in\mathcal H_{\mathrm{in}}}
\|\mu-\mu^*\|_{L^2(P_{X,\widetilde Y})}^2}_{\text{first-step approximation error}}+4\underbrace{\inf_{f\in\mathcal H_{\mathrm{out},\lambda}}
\|f-f^*\|_{L^2(P_X)}^2}_{\text{second-step approximation error}}\\
&+6(1-\lambda)^2\underbrace{\mathbb E\left[
\|\widehat\mu-\mu^*\|_{L^2(P_{X,\widetilde Y})}^2
-\inf_{\mu\in\mathcal H_{\mathrm{in}}}\|\mu-\mu^*\|_{L^2(P_{X,\widetilde Y})}^2
\right]}_{\text{first-step statistical error}}\\
&+2\underbrace{\mathbb E\left[
\|\widehat f_\lambda-g_{\lambda,\widehat\mu}\|_{L^2(P_X)}^2
-\inf_{f\in\mathcal H_{\mathrm{out},\lambda}}\|f-g_{\lambda,\widehat\mu}\|_{L^2(P_X)}^2
\right]}_{\text{second-step statistical error}}.
\end{align*}
At \(\lambda=1\), the calibration terms are omitted and \(g_{1,\widehat\mu}=f^*\). 

Define the second-step sample-size scale
\begin{equation}
N_\lambda=\left\{\frac{\lambda}{N_\ell}\vee
\frac{1-\lambda}{N_u}\right\}^{-1}.
\label{eq:weighted-size}
\end{equation}
It satisfies \(N_0=N_u\), \(N_{N_\ell/N}=N\), and \(N_1=N_\ell\). Combining ReLU approximation bounds with least-squares oracle inequalities yields the following risk bound.

\begin{theorem}[Approximation and statistical errors]
\label{thm:weighted-holder-risk}
Suppose Assumptions~\ref{ass:holder-smoothness}
and~\ref{ass:subgaussian-noise} hold, with \(N_\ell,N_u\ge2\).
For deterministic \(\lambda\in[0,1]\), set
\(n_{\mathrm{in}}=N_\ell\) and \(n_{\mathrm{out}}=N_\lambda\).
For \(a\in\{\mathrm{in},\mathrm{out}\}\), choose integers
\(2\le K_a\le\lfloor n_a\rfloor\).
There exist network classes from Definition~\ref{def:relu-network},
used as in \eqref{eq:nn-working-classes}, with fixed clipping levels
\(U_a=1\vee B_a\) and parameter orders
\(L_a\asymp\log(2K_a)\), \(W_a\asymp K_a\),
\(S_a\asymp K_a\log(2K_a)\), and \(R_a=O(B_a\vee1)\),
such that
\begin{align*}
\mathcal R_\lambda
\le C\biggl\{
(1-\lambda)^2B_{\mathrm{in}}^2K_{\mathrm{in}}^{-2{\beta}_{\mathrm{in}}/d_{\mathrm{in}}}
+B_{\mathrm{out}}^2K_{\mathrm{out}}^{-2{\beta}_{\mathrm{out}}/d_x}
+(1-\lambda)^2\frac{K_{\mathrm{in}}\log^3(2N_{\ell})}{N_{\ell}}
+\frac{K_{\mathrm{out}}\log^3(2{N}_\lambda)}{{N}_\lambda}\biggr\}.
\end{align*}
The constant \(C\) depends only on the fixed class parameters, not on \(\lambda\).
\end{theorem}

Balancing approximation and statistical errors, including logarithmic factors, yields the following corollary.

\begin{corollary}[Convergence rate]
\label{cor:weighted-convergence-rate}
Under the conditions of Theorem~\ref{thm:weighted-holder-risk}, choose \(K_a\asymp\{n_a/\log^3(2n_a)\}^{d_a/(2{\beta}_a+d_a)}\) for \(a\in\{\mathrm{in},\mathrm{out}\}\), rounded and truncated to admissible integers in \([2,\lfloor n_a\rfloor]\).
Define
\[
r_{\mathrm{in}}=\frac{2{\beta}_{\mathrm{in}}}{2{\beta}_{\mathrm{in}}+d_x+1},
\qquad
r_{\mathrm{out}}=\frac{2{\beta}_{\mathrm{out}}}{2{\beta}_{\mathrm{out}}+d_x}.
\]
Equivalently, $r_a=2{\beta}_a/(2{\beta}_a+d_a)$. Then
\begin{equation*}
\mathcal R_\lambda
\lesssim (1-\lambda)^2N_{\ell}^{-r_{\mathrm{in}}}\log^{3r_{\mathrm{in}}}(2N_{\ell})+{N}_\lambda^{-r_{\mathrm{out}}}\log^{3r_{\mathrm{out}}}(2{N}_\lambda).
\end{equation*}
\end{corollary}
The two terms correspond to calibration error from labeled data, scaled by \((1-\lambda)^2\), and second-step regression error, respectively.
Table~\ref{tab:special-weight-risk-bounds} summarizes the bounds for three special weights. To compare their convergence rates, suppose \(N_u\asymp N_\ell^\kappa\) with \(\kappa>1\). For \(\lambda=O(N_\ell/N_u)\), which includes the first two weights in Table~\ref{tab:special-weight-risk-bounds}, the wPPRC bound has polynomial exponent \(\min\{r_{\mathrm{in}},\kappa r_{\mathrm{out}}\}\), compared with \(r_{\mathrm{out}}\) for supervised regression. Thus, \(r_{\mathrm{in}}>r_{\mathrm{out}}\) yields a strictly faster rate, even with the logarithmic factors. Calibration error determines the polynomial rate once
\(\kappa r_{\mathrm{out}}\ge r_{\mathrm{in}}\).

\begin{table}[htbp]
\centering
\caption{Risk upper bounds for wPPRC at three special weights.}
\label{tab:special-weight-risk-bounds}
\small
\begin{tabular}{@{}lll@{}}
\toprule
Estimator & Weight \(\lambda\) & Risk bound, up to a fixed constant \\
\midrule
\(\widehat f_0\) & \(0\) &
\(N_\ell^{-r_{\mathrm{in}}}\log^{3r_{\mathrm{in}}}(2N_\ell)+N_u^{-r_{\mathrm{out}}}\log^{3r_{\mathrm{out}}}(2N_u)\) \\
\(\widehat f_{N_\ell/N}\) & \(N_\ell/N\) &
\((N_u/N)^2N_\ell^{-r_{\mathrm{in}}}\log^{3r_{\mathrm{in}}}(2N_\ell)+N^{-r_{\mathrm{out}}}\log^{3r_{\mathrm{out}}}(2N)\) \\
\(\widehat f_1\) (Supervised) & \(1\) &
\(N_\ell^{-r_{\mathrm{out}}}\log^{3r_{\mathrm{out}}}(2N_\ell)\) \\
\bottomrule
\end{tabular}
\end{table}

The condition \(r_{\mathrm{in}}>r_{\mathrm{out}}\) requires \(\beta_{\mathrm{in}}>\beta_{\mathrm{out}}(1+1/d_x)\), so calibration must be sufficiently smoother to offset its additional input dimension. This is possible because
\(f^*(x)=\int\mu^*(x,\widetilde y)\,dP_{\widetilde Y\mid X=x}(\widetilde y)\)
also depends on how the conditional distribution of the pseudo-label varies with \(x\). Exploiting structure in the calibration function can relax this smoothness requirement. For example, consider the partially linear model
\begin{equation}
\mu^*(x,\widetilde y)=h_0(x)+a_0\widetilde y.
\label{eq:partially-linear-structure}
\end{equation}
Since the linear term can be represented exactly by a ReLU network, only \(h_0\) requires nonparametric approximation. The following proposition gives the resulting risk bound.

\begin{proposition}[Partially Linear Calibration]
\label{prop:partially-linear-rate}
Let \(N_\ell,N_u\ge2\) and \(\lambda\in[0,1]\) be deterministic. Suppose \eqref{eq:partially-linear-structure} holds on \(\Omega_{\mathrm{in}}\), where \(h_0\in C^{\beta_{\mathrm{in}}}(\Omega_{\mathrm{out}},B_h)\) and \(|a_0|\le A_0\) for fixed \(B_h,A_0>0\). Assume the smoothness condition on \(f^*\) in Assumption~\ref{ass:holder-smoothness} and the noise condition in Assumption~\ref{ass:subgaussian-noise}. Set \(r_{\mathrm{in},x}=2\beta_{\mathrm{in}}/(2\beta_{\mathrm{in}}+d_x)\). Choose \(K_{\mathrm{in}}\asymp N_\ell^{d_x/(2\beta_{\mathrm{in}}+d_x)}\) and \(K_{\mathrm{out}}\asymp N_\lambda^{d_x/(2\beta_{\mathrm{out}}+d_x)}\), rounded and truncated to integers in \([2,N_\ell]\) and \([2,\lfloor N_\lambda\rfloor]\), respectively. There exist network classes of the form \eqref{eq:nn-working-classes}, with fixed clipping levels \(U_{\mathrm{in}}\ge1\vee\|\mu^*\|_\infty\) and \(U_{\mathrm{out}}\ge1\vee\|f^*\|_\infty\), and parameter orders \(L_a\asymp\log(2K_a)\), \(W_a\asymp K_a\), \(S_a\asymp K_a\log(2K_a)\) for \(a\in\{\mathrm{in},\mathrm{out}\}\), \(R_{\mathrm{in}}=O(B_h\vee A_0\vee1)\), and \(R_{\mathrm{out}}=O(B_{\mathrm{out}}\vee1)\), such that
\begin{equation*}
\mathcal R_\lambda\lesssim
(1-\lambda)^2N_\ell^{-r_{\mathrm{in},x}}\log^3(2N_\ell)
+N_\lambda^{-r_{\mathrm{out}}}\log^3(2N_\lambda).
\end{equation*}
The implicit constant depends only on the fixed class parameters, not on \(\lambda\).
\end{proposition}

Here, \(r_{\mathrm{in},x}>r_{\mathrm{out}}\) is equivalent to \(\beta_{\mathrm{in}}>\beta_{\mathrm{out}}\). More calibration examples are given in Section~S3.3 of the SM.

\subsection{Risk Bound for the VS Estimator}
We select \(\widehat\lambda\) by validation as described in Section~\ref{sec:weight-selection}. The following proposition gives a risk guarantee for the VS estimator.

\begin{proposition}[Risk bound for the VS estimator]
\label{prop:validation-selection-optimized}
Suppose the conditions of Corollary~\ref{cor:weighted-convergence-rate} hold, and construct the candidates \(\widehat f_\lambda\), \(\lambda\in\Lambda\). Then 
\begin{itemize}[topsep=0pt, partopsep=0pt, before={\predisplaypenalty=0}]
\item[(i)] The VS estimator satisfies
\begin{equation*}
\mathcal R_{\widehat\lambda}
\le\min_{\lambda\in\Lambda}\left\{
\mathcal R_\lambda+C\sqrt{\frac{\mathcal R_\lambda\{\log(2|\Lambda|)+1\}}{N_{\ell,\mathrm{val}}}}
\right\}
+C\frac{\log(2|\Lambda|)+1}{N_{\ell,\mathrm{val}}}.
\end{equation*}
\item[(ii)] If \(\{0,1\}\subset\Lambda\), \(|\Lambda|\) is bounded, and \(N_{\ell,\mathrm{val}}\asymp N_\ell\), then
\begin{equation*}
\mathcal R_{\widehat\lambda}
\le\min\{\mathcal R_0,\mathcal R_1\}
+C\sqrt{\frac{\min\{\mathcal R_0,\mathcal R_1\}}{N_\ell}}
+\frac{C}{N_\ell}.
\end{equation*}
\end{itemize}
\end{proposition}

Part (i) compares the expected risk of the VS estimator with the smallest expected risk among the candidates and quantifies the cost of selection based on a finite validation sample. When \(|\Lambda|\) is bounded and \(N_{\ell,\mathrm{val}}\asymp N_\ell\), the validation remainder is of smaller order than the risk bounds in Corollary~\ref{cor:weighted-convergence-rate}.
For example, part (ii) considers a candidate set containing both \(\lambda=0\) and \(\lambda=1\).
Using the risk bound for the supervised estimator, the remainder is
\(O\{N_\ell^{-(1+r_{\mathrm{out}})/2}\log^{3r_{\mathrm{out}}/2}(2N_\ell)+N_\ell^{-1}\}
=o\{N_\ell^{-r_{\mathrm{out}}}\log^{3r_{\mathrm{out}}}(2N_\ell)\}\).
Using the risk bound for \(\widehat f_0\), it is
\(O\{N_\ell^{-(1+r_{\mathrm{in}})/2}\log^{3r_{\mathrm{in}}/2}(2N_\ell)
+N_\ell^{-1/2}N_u^{-r_{\mathrm{out}}/2}\log^{3r_{\mathrm{out}}/2}(2N_u)+N_\ell^{-1}\}
=o\{N_\ell^{-r_{\mathrm{in}}}\log^{3r_{\mathrm{in}}}(2N_\ell)+N_u^{-r_{\mathrm{out}}}\log^{3r_{\mathrm{out}}}(2N_u)\}\).
Thus, validation selection preserves the faster convergence rate among the risk bounds for the two candidate estimators. 
Since \(N_{\ell,\mathrm{val}}\asymp N_\ell\), these rates also hold in terms of the total labeled sample size \(N_\ell+N_{\ell,\mathrm{val}}\).

\section{Minimax Bounds and Comparisons}
\label{sec:minimax-bounds}
In this section, we establish minimax lower bounds and identify when
wPPRC is minimax optimal up to logarithmic factors. We also compare minimax
risks across observation structures and examine whether validation
selection preserves the gains from pseudo-labels.

Fix \({\beta}_a,B_a,\Omega_a\), \(a\in\{\mathrm{in},\mathrm{out}\}\), and \(\sigma\), and let
\[\mathcal P=\mathcal P(({\beta}_a,B_a,\Omega_a)_{a\in\{\mathrm{in},\mathrm{out}\}},\sigma)\] denote the class of joint distributions \(P=P_{X,Y,\widetilde Y}\) whose conditional means
$\mu^*_P(x,\tilde y)=\mathbb{E}_P(Y\mid X=x,\widetilde Y=\tilde y)$ and
$f^*_P(x)=\mathbb{E}_P(Y\mid X=x)$ satisfy
Assumption~\ref{ass:holder-smoothness}, and whose responses satisfy Assumption~\ref{ass:subgaussian-noise}.
The minimax risk for estimating \(f^*\) from
\(\mathcal L\) and \(\mathcal U\) is
\[
\mathfrak M(N_{\ell},N_u;\mathcal P)
=
\inf_{\widehat f}\,
\sup_{P\in\mathcal P}\,
\mathbb{E}_P\,
\|\widehat f-f_P^*\|_{L^2(P_X)}^2 .
\]
Here the infimum is over all measurable estimators based on
\((\mathcal L,\mathcal U)\).
Define
\[
r_* = r_{\mathrm{in}}\vee r_{\mathrm{out}}.
\]
The exponent \(r_*\) reflects the joint smoothness restrictions on \(\mu^*\) and \(f^*\) and governs the labeled-sample term in the following minimax lower bound.

\begin{theorem}[Minimax lower bound]
\label{thm:minimax-lower-main}
For the class \(\mathcal P\) defined above, there is a constant \(c>0\),
depending only on the fixed class parameters, such that for
\(N_\ell\ge1\) and \(N_u\ge0\),
\begin{equation*}
\mathfrak M(N_\ell,N_u;\mathcal P)
\ge c\left\{
N_\ell^{-r_*}+(N_\ell+N_u)^{-r_{\mathrm{out}}}
\right\}.
\end{equation*}
\end{theorem}

The two terms reflect the information available in the labeled and
unlabeled samples. Write \(P_{X,\widetilde Y,Y}
=P_X\,P_{\widetilde Y\mid X}\,P_{Y\mid X,\widetilde Y}\) and 
\(P_{X,\widetilde Y}
=P_X\,P_{\widetilde Y\mid X}.
\)
Both samples contain \(X\) and \(\widetilde Y\), whereas only \(\mathcal L\) contains the true response \(Y\). 
For the first term, consider a subclass of distributions that share the same marginal law of \((X,\widetilde Y)\) but have different regression functions \(f^*\). Only the labeled observations can distinguish these regression functions. Under the joint smoothness constraints on \(\mu^*\) and \(f^*\), this yields the term \(N_\ell^{-r_*}\). For the second term, consider a subclass of distributions sharing the same affine calibration function \(\mu^*(x,\tilde y)=b+a\tilde y\), with \(a\ne0\). The target satisfies \(f^*(x)=b+a\mathbb E(\widetilde Y\mid X=x)\). With this calibration function fixed, estimating \(f^*\) is equivalent to estimating \(\mathbb E(\widetilde Y\mid X)\). Both samples provide covariate--pseudo-label pairs, yielding the term \((N_\ell+N_u)^{-r_{\mathrm{out}}}\).

\subsection{Minimax Optimality}
\label{sec:minimax-optimality}

We compare the risk upper bounds in Corollary~\ref{cor:weighted-convergence-rate} with the minimax lower bound in Theorem~\ref{thm:minimax-lower-main} to identify when the wPPRC estimators are minimax optimal up to logarithmic factors. Table~\ref{tab:upper-lower-rate-comparison} summarizes this comparison when \(N_u\asymp N_\ell^\kappa\), for fixed \(\kappa>1\). For deterministic \(\lambda=O(N_\ell/N_u)\), including \(\lambda=0\) and \(\lambda=N_\ell/N\), the risk upper bound for wPPRC matches the minimax lower bound up to logarithmic factors when \(r_{\mathrm{in}}\ge r_{\mathrm{out}}\). When \(r_{\mathrm{in}}<r_{\mathrm{out}}\), this bound no longer matches the lower bound, but the supervised estimator $\widehat{f}_1$ attains the minimax rate up to logarithmic factors.

\begin{table}[htbp]
\centering
\caption{Comparison of polynomial rates. Logarithmic factors are omitted; bold entries indicate matching upper and lower bounds.}
\label{tab:upper-lower-rate-comparison}
\small
\setlength{\tabcolsep}{4pt}
\begin{tabular}{@{}lccc@{}}
\toprule
Regime & \begin{tabular}{@{}c@{}}Risk bound for wPPRC\\\(\lambda=O(N_\ell/N_u)\)\end{tabular}
& \begin{tabular}{@{}c@{}}Risk bound for wPPRC\\\(\lambda=1\) (supervised)\end{tabular}
& Minimax lower bound \\
\midrule
\(r_{\mathrm{in}}\ge r_{\mathrm{out}}\)
& {\boldmath\(N_\ell^{-\min\{r_{\mathrm{in}},\kappa r_{\mathrm{out}}\}}\)}
& \(N_\ell^{-r_{\mathrm{out}}}\)
& {\boldmath\(N_\ell^{-\min\{r_{\mathrm{in}},\kappa r_{\mathrm{out}}\}}\)} \\
\(r_{\mathrm{in}}<r_{\mathrm{out}}\)
& \(N_\ell^{-r_{\mathrm{in}}}\)
& {\boldmath\(N_\ell^{-r_{\mathrm{out}}}\)}
& {\boldmath\(N_\ell^{-r_{\mathrm{out}}}\)} \\
\bottomrule
\end{tabular}
\end{table}

\subsection{Minimax Comparisons Across Observation Structures}

We compare four baseline observation structures: supervised (Sup), semi-supervised (SS), supervised with pseudo-labels (Sup-P), and augmented supervised (Aug). All settings share the same target \(f_P^*\) and model class \(\mathcal P\), but differ in the observed variables, as summarized in Table~\ref{tab:reference-lower-bounds}. Here, \(\mathcal L_{XY}=\{(X_i,Y_i)\}_{i=1}^{N_\ell}\) and \(\mathcal U_X=\{X_j\}_{j=1}^{N_u}\) denote the labeled pairs and additional covariates, respectively.

For each setting \(\iota\), define its minimax risk by
\[
\mathfrak M_\iota(N_\ell,N_u;\mathcal P)
=
\inf_{\widehat f}\sup_{P\in\mathcal P}
\mathbb E_P\|\widehat f-f_P^*\|_{L^2(P_X)}^2,
\]
where the infimum is over estimators using only the observations available in that setting. We retain \(N_u\) for notational consistency, although the Sup and Sup-P do not depend on it. 

\begin{table}[!htbp]
\centering
\caption{Minimax lower bounds under different observation structures on the class \(\mathcal P\).
$\checkmark$ indicates an observed variable, and -- an unobserved variable.}
\label{tab:reference-lower-bounds}
\begingroup
\small
\setlength{\tabcolsep}{3pt}
\renewcommand{\arraystretch}{1.2}
\begin{tabular}{@{}ll*{3}{>{\centering\arraybackslash}m{1.25em}}@{\hspace{9pt}}*{3}{>{\centering\arraybackslash}m{1.25em}}l@{}}
\toprule
\multirow{2}{*}{Setting}
& \multirow{2}{*}{Observations}
& \multicolumn{3}{c}{Labeled}
& \multicolumn{3}{c}{Unlabeled}
& \multirow{2}{*}{Lower bound} \\
\cmidrule(lr){3-5}\cmidrule(lr){6-8}
& & $X$ & $Y$ & $\widetilde Y$
& $X$ & $Y$ & $\widetilde Y$ & \\
\midrule
Our setting & $\mathcal L,\mathcal U$
& \cellcolor{black!8}$\checkmark$ & \cellcolor{black!8}$\checkmark$ & \cellcolor{black!8}$\checkmark$
& \cellcolor{black!8}$\checkmark$ & -- & \cellcolor{black!8}$\checkmark$
& $N_{\ell}^{-r_*}+(N_{\ell}+N_u)^{-r_{\mathrm{out}}}$ \\
Sup & $\mathcal L_{XY}$
& \cellcolor{black!8}$\checkmark$ & \cellcolor{black!8}$\checkmark$ & --
& -- & -- & --
& $N_{\ell}^{-r_{\mathrm{out}}}$ \\
SS & $\mathcal L_{XY},\mathcal U_X$
& \cellcolor{black!8}$\checkmark$ & \cellcolor{black!8}$\checkmark$ & --
& \cellcolor{black!8}$\checkmark$ & -- & --
& $N_{\ell}^{-r_{\mathrm{out}}}$ \\
Sup-P & $\mathcal L$
& \cellcolor{black!8}$\checkmark$ & \cellcolor{black!8}$\checkmark$ & \cellcolor{black!8}$\checkmark$
& -- & -- & --
& $N_{\ell}^{-r_{\mathrm{out}}}$ \\
Aug & $\mathcal L,\mathcal U_X$
& \cellcolor{black!8}$\checkmark$ & \cellcolor{black!8}$\checkmark$ & \cellcolor{black!8}$\checkmark$
& \cellcolor{black!8}$\checkmark$ & -- & --
& $N_{\ell}^{-r_{\mathrm{out}}}$ \\
\bottomrule
\end{tabular}
\endgroup
\end{table}

All four baselines have a minimax lower bound of order \(N_\ell^{-r_{\mathrm{out}}}\), matching the risk bound for the supervised estimator up to logarithmic factors. Comparing these lower bounds with the risk upper bound for wPPRC gives the following result.

\begin{theorem}[Strict improvement over baseline minimax risks]
\label{thm:baseline-minimax-separation}
Under the conditions and network parameter choices of Corollary~\ref{cor:weighted-convergence-rate}, let \(\lambda=O(N_\ell/N_u)\) be deterministic.
If $r_{\mathrm{in}}>r_{\mathrm{out}}$ and $N_u\asymp N_{\ell}^\kappa$ for fixed
$\kappa>1$, then, as $N_{\ell}\to\infty$,
\[
\frac{\sup_{P\in\mathcal P}\mathbb E_P\|\widehat f_{\lambda}-f_P^*\|_{L^2(P_X)}^2}
{\mathfrak M_\iota(N_{\ell},N_u;\mathcal P)}\longrightarrow0,
\qquad \iota\in\{\mathrm{Sup},\mathrm{SS},\mathrm{Sup\text{-}P},\mathrm{Aug}\}.
\]
\end{theorem}

Theorem \ref{thm:baseline-minimax-separation} shows that wPPRC can achieve a worst-case risk of smaller order than the minimax risks in all four baseline settings. In particular, Aug observes the same labeled triples and additional covariates, but lacks pseudo-labels on the additional sample. The improvement over this baseline demonstrates the value of additional pseudo-labels.

\begin{remark}[Generator access]
Pseudo-labels may be unavailable on the additional sample when the generator or its required inputs are inaccessible. For example, if \(\widetilde Y=g(X,Z)\), knowing \(g\) and observing \(X\) may not suffice when \(Z\) is unavailable. If the generator and all required inputs are available, these pseudo-labels can be obtained.
\end{remark}

\begin{remark}[Weaker sample growth conditions]
The conclusion of Theorem~\ref{thm:baseline-minimax-separation} also holds under the weaker condition \((N_\ell/N_u)^{r_{\mathrm{out}}}\log^{3r_{\mathrm{out}}}(2N_u)\to0\), with the other conditions unchanged. For example, \(N_u\asymp N_\ell(\log N_\ell)^{\kappa_{\log}}\) with \(\kappa_{\log}>3\) suffices.
\end{remark}

\subsection{Minimax Comparisons Under Validation Selection}
We now show that validation selection preserves the improvement in Theorem~\ref{thm:baseline-minimax-separation}.

\begin{corollary}
\label{cor:validation-minimax-separation}
Under the conditions of Theorem~\ref{thm:baseline-minimax-separation} and Proposition~\ref{prop:validation-selection-optimized}, suppose that \(|\Lambda|\) is bounded, \(N_{\ell,\mathrm{val}}\asymp N_\ell\), and \(\Lambda\) contains a deterministic weight \(\lambda=O(N_\ell/N_u)\). Then, as \(N_\ell\to\infty\),
\begin{equation*}
\frac{\sup_{P\in\mathcal P}\mathbb E_P
\|\widehat f_{\widehat\lambda}-f_P^*\|_{L^2(P_X)}^2}
{\mathfrak M_\iota(N_\ell+N_{\ell,\mathrm{val}},N_u;\mathcal P)}
\longrightarrow0,
\qquad \iota\in\{\mathrm{Sup},\mathrm{SS},\mathrm{Sup\text{-}P},\mathrm{Aug}\}.
\end{equation*}
\end{corollary}

Corollary~\ref{cor:validation-minimax-separation} shows that validation selection retains the gains from additional pseudo-labels, even when the baseline estimators receive the same total number of labeled observations. Including \(0\) or \(N_\ell/N\) in the candidate set satisfies the required weight condition.

\section{Linear Regression and Variance-Optimal Weighting}
\label{sec:linear-regression-weighting}
We consider a linear special case with explicit solutions to illustrate how calibration uses pseudo-labels and how \(\lambda\) affects estimation variance. Suppose \(f^*(x)=x^\top\vartheta_0\) and \(\mu^*(x,\widetilde y)=x^\top b_0+a_0\widetilde y\). 
In this setting, estimating \(f^*\) reduces to estimating \(\vartheta_0\), with the fitted predictor given by \(\widehat f_\lambda(x)=x^\top\widehat\vartheta_\lambda\). Let \(\alpha_0=\arg\min_{\alpha}\mathbb E(\widetilde Y-X^\top\alpha)^2=\{\mathbb E(XX^\top)\}^{-1}\mathbb E(X\widetilde Y)\) denote the population linear projection coefficient of \(\widetilde Y\) on \(X\), where \(\mathbb E(XX^\top)\) is nonsingular. The conditional-mean identity \(f^*(X)=\mathbb E\{\mu^*(X,\widetilde Y)\mid X\}\) then implies \(\vartheta_0=b_0+a_0\alpha_0\).

We fit linear models by ordinary least squares (OLS) in both steps.
Let \(X_\ell\) and \(X_u\) be the design matrices for the labeled and unlabeled samples, respectively, with response vector \(y_\ell\) and pseudo-label vectors \(\widetilde y_\ell,\widetilde y_u\).
The following expressions hold whenever the relevant sample matrices are nonsingular.
\begin{itemize}[leftmargin=1.5em, labelsep=0.5em, topsep=2pt, before={\predisplaypenalty=0}]
\item \textbf{Step 1: Calibration with \((X,\widetilde Y)\).} Regress \(Y\) on \((X,\widetilde Y)\) using \(\mathcal L\) to obtain 
\begin{align*}
(\widehat b^\top,\widehat a)^\top
&=\arg\min_{b,a}\frac{1}{N_\ell}\|y_\ell-X_\ell b-a\widetilde y_\ell\|_2^2
=\left([X_\ell,\widetilde y_\ell]^\top[X_\ell,\widetilde y_\ell]\right)^{-1}[X_\ell,\widetilde y_\ell]^\top y_\ell\\
&=\left(
(y_\ell-\widehat a\widetilde y_\ell)^\top X_\ell(X_\ell^\top X_\ell)^{-1},
\quad
\frac{\widetilde y_\ell^\top\{I-X_\ell(X_\ell^\top X_\ell)^{-1}X_\ell^\top\}y_\ell}
{\widetilde y_\ell^\top\{I-X_\ell(X_\ell^\top X_\ell)^{-1}X_\ell^\top\}\widetilde y_\ell}
\right)^\top.
\end{align*}
\item \textbf{Step 2: Weighted regression on \(X\).} Regress observed responses and calibrated pseudo-labels on \(X\): 
\begin{align*}
\widehat\vartheta_\lambda
&=\arg\min_\vartheta\left\{
\frac{1-\lambda}{N_u}\|X_u\widehat b+\widehat a\widetilde y_u-X_u\vartheta\|_2^2
+\frac{\lambda}{N_\ell}\|y_\ell-X_\ell\vartheta\|_2^2
\right\}\\
&=\left(\frac{1-\lambda}{N_u}X_u^\top X_u+\frac{\lambda}{N_\ell}X_\ell^\top X_\ell\right)^{-1}
\left\{\frac{1-\lambda}{N_u}X_u^\top(X_u\widehat b+\widehat a\widetilde y_u)
+\frac{\lambda}{N_\ell}X_\ell^\top y_\ell\right\}\\
&=\widehat b+\widehat a\,\widehat\alpha_\lambda,
\end{align*}
where \(\widehat\alpha_\lambda\) is the weighted least-squares estimate of \(\alpha_0\):
\begin{align*}
\widehat\alpha_\lambda
&=\arg\min_\alpha\left\{
\frac{1-\lambda}{N_u}\|\widetilde y_u-X_u\alpha\|_2^2
+\frac{\lambda}{N_\ell}\|\widetilde y_\ell-X_\ell\alpha\|_2^2
\right\}\\
&=\left(\frac{1-\lambda}{N_u}X_u^\top X_u+\frac{\lambda}{N_\ell}X_\ell^\top X_\ell\right)^{-1}
\left(\frac{1-\lambda}{N_u}X_u^\top\widetilde y_u+\frac{\lambda}{N_\ell}X_\ell^\top\widetilde y_\ell\right).
\end{align*}
\end{itemize}

The coefficient \(\widehat a\) captures the contribution of the pseudo-label beyond \(X\), while \(X^\top\widehat b\) captures the remaining linear dependence on \(X\). The nonparametric calibration function \(\widehat\mu\) allows nonlinear dependence on both inputs. The predictor \(x^\top\widehat\vartheta_\lambda=x^\top(\widehat b+\widehat a\widehat\alpha_\lambda)\) uses only \(X\) at deployment. Even when \(\lambda=0\), labeled responses contribute through \(\widehat b\) and \(\widehat a\).

\smallskip
\noindent \textbf{Leading Covariance and Optimal Weighting.}
Recall that \(\varepsilon=Y-\mu^*(X,\widetilde Y)\), and let \(\eta=\mu^*(X,\widetilde Y)-f^*(X)\). The decomposition \(Y-f^*(X)=\eta+\varepsilon\) separates the response variation explained by the pseudo-label beyond \(X\) from the remaining noise. Under the regularity conditions stated in the SM, the leading covariance of \(\widehat\vartheta_\lambda\) for deterministic \(\lambda\) is
\begin{align*}
\Sigma_\lambda
&=\{\mathbb E(XX^\top)\}^{-1}
\left[
\frac{\mathbb E(XX^\top\varepsilon^2)}{N_\ell}
+\left\{\frac{\lambda^2}{N_\ell}+\frac{(1-\lambda)^2}{N_u}\right\}
\mathbb E(XX^\top\eta^2)
\right]
\{\mathbb E(XX^\top)\}^{-1}.
\end{align*}
The first term depends only on the labeled sample size and is unaffected by \(\lambda\). Minimizing \(\lambda^2/N_\ell+(1-\lambda)^2/N_u\) gives \(\lambda=N_\ell/(N_\ell+N_u)\), which minimizes the leading covariance within the two-step OLS family. Relative to supervised OLS (\(\lambda=1\)), the reduction is
\begin{align*}
\Sigma_1-\Sigma_{N_\ell/(N_\ell+N_u)}
=\frac{N_u}{N_\ell(N_\ell+N_u)}
\{\mathbb E(XX^\top)\}^{-1}
\mathbb E(XX^\top\eta^2)
\{\mathbb E(XX^\top)\}^{-1}.
\end{align*}
At \(x\), the corresponding reduction in leading prediction variance is \(x^\top\{\Sigma_1-\Sigma_{N_\ell/(N_\ell+N_u)}\}x\ge0\). The ReLU estimator with the same weight attains the nonparametric minimax rate up to logarithmic factors when \(r_{\mathrm{in}}\ge r_{\mathrm{out}}\) (Section~\ref{sec:minimax-optimality}). For general nonparametric regression, validation selects \(\lambda\) with the risk guarantee in Proposition~\ref{prop:validation-selection-optimized}.

\section{Connections With Existing Methods}
\label{sec:connections-existing-methods}

We relate wPPRC to existing methods that use pseudo-labels or auxiliary information, comparing how they incorporate this information, when their objectives or estimators coincide with those of wPPRC, and how their theoretical guarantees differ. These comparisons place calibration and weighting in the context of existing approaches to prediction and inference.

\noindent \textbf{Prediction Aided by Surrogate Training (PAST).}
With the same calibration fit and output class, the squared-loss objective of PAST \citep{xia2024prediction} coincides with that of wPPRC at \(\lambda=N_\ell/N\). The PAST analysis establishes oracle inequalities for general function classes and losses. It examines when auxiliary information helps or harms prediction and when the estimator attains the accuracy of a fully labeled oracle. We extend this construction to a weighted family, establish minimax comparisons across observation structures, and give an expected-risk guarantee for validation-based weight selection.

\smallskip
\noindent \textbf{Coupled Training.} \citet{shi2026coupled} jointly train a deployment predictor using \(X\) and an auxiliary predictor using privileged information, encouraging agreement on the additional sample. In wPPRC, calibration is fitted first and held fixed during the second-step regression.

\smallskip
\noindent \textbf{Pseudo-Labeling and PPI.}
Direct pseudo-labeling (PL) treats pseudo-labels as additional responses, generally targeting a weighted average of \(f^*(X)\) and \(\mathbb E(\widetilde Y\mid X)\). PPI \citep{angelopoulos2023prediction} and PPI++ \citep{angelopoulos2023ppi++} use labeled observations to correct the pseudo-label loss so that the expected objective equals the risk based on the true response. 
For flexible predictor classes, however, this empirical correction can favor predictions near the allowed output boundaries and destabilize minimization. The wPPRC objective is a nonnegative weighted sum of squared losses, avoiding this loss-subtraction mechanism.

\smallskip
\noindent \textbf{SynSurr.} \citet{mccaw2024synthetic} develop SynSurr to estimate genetic effects using partially observed phenotypes and synthetic surrogates. When both methods use the same labeled and unlabeled samples and the same intercept specification, the SynSurr coefficient estimate equals the linear wPPRC estimate in Section~\ref{sec:linear-regression-weighting} at \(\lambda=N_\ell/(N_\ell+N_u)\). Thus, wPPRC extends this two-step regression construction to nonparametric prediction.

Section~S6 of the SM provides detailed objectives and derivations for these comparisons.

\section{Empirical Studies}
\label{sec:empirical-studies}
We evaluate the performance of the proposed estimators through simulations and an application to homeless-count prediction. Additional results and implementation details are provided in Section~S7 of the SM.

\subsection{Simulations}
\label{sec:simulation-studies}

The simulation study uses \(B=100\) replications to compare regression accuracy across pseudo-label mechanisms and examine the effect of weight selection. In each replication, 1,000 labeled observations are randomly split into training (\(N_\ell=700\)), tuning (\(N_{\ell,\mathrm{tune}}=150\)), and validation (\(N_{\ell,\mathrm{val}}=150\)) subsets. The tuning and validation subsets are used for checkpoint and weight selection, respectively. The unlabeled sample contains \(N_u=10{,}000\) observations, and an independent test sample contains \(N_{\mathrm{test}}=10{,}000\) observations.

We compare three fixed-weight estimators from the wPPRC family: wPPRC(0), wPPRC \allowbreak(\(N_\ell/N\)), and supervised regression (Sup), corresponding to \(\lambda=0\), \(N_\ell/N\), and \(1\), respectively. We also evaluate the validation-selected estimator (VS), 
which selects among candidates fitted with weights in  \(\Lambda=\{0,\allowbreak 0.05,\allowbreak N_\ell/N,\allowbreak 0.1,\allowbreak 0.25,\allowbreak 0.5,\allowbreak 0.75,\allowbreak 0.9,\allowbreak 1\}\). PL and PPI are included as additional baselines, with \(\lambda_{\mathrm{pl}}=N_u/N_\ell\) and \(\lambda_{\mathrm{ppi}}=1\), respectively.

\smallskip
\noindent \textbf{Evaluation metrics.}  For each method \(m\in\{\mathrm{Sup},\allowbreak \text{wPPRC}(0),\allowbreak \text{wPPRC}(N_\ell/N),\allowbreak \mathrm{VS},\allowbreak \mathrm{PL},\allowbreak \mathrm{PPI}\}\), let \(\widehat f_m\) denote its fitted estimator. Prediction risk and excess risk are evaluated on an independent test sample:
\[
\widehat R_m=\frac{1}{N_{\mathrm{test}}}\sum_{k=1}^{N_{\mathrm{test}}}
  \{Y_k-\widehat f_m(X_k)\}^2,\qquad
\widehat{\mathcal E}_m=\frac{1}{N_{\mathrm{test}}}\sum_{k=1}^{N_{\mathrm{test}}}
  \{\widehat f_m(X_k)-f^*(X_k)\}^2.
\]
The corresponding relative improvements over the supervised estimator are:
\[
\mathit{RIR}_m=100\%\times
  \frac{\widehat R_{\mathrm{Sup}}-\widehat R_m}{\widehat R_{\mathrm{Sup}}},
\qquad
\mathit{RI}_m=100\%\times
  \frac{\widehat{\mathcal E}_{\mathrm{Sup}}-\widehat{\mathcal E}_m}
       {\widehat{\mathcal E}_{\mathrm{Sup}}}.
\]
Positive values indicate lower risk than that of the supervised estimator. Each metric is computed within each replication and summarized by its Monte Carlo mean and standard error across the \(B\) replications.

\smallskip
\noindent \textbf{Data-generating processes.} We set \(d_x=d_t=100\) and generate $X\sim \mathrm{N}(0,\Sigma_X)$, $T\sim \mathrm{N}(0,I_{d_t})$, $X\perp T$,
where \((\Sigma_X)_{jj}=1\) and \((\Sigma_X)_{jk}=0.1\) for \(j\ne k\).
Let
$\operatorname{expit}(u)=1/(1+\exp(-u))$,
$b=5^{-1/2}(1,1,1,1,1,0,\ldots,0)^\top$.
The response is generated as
\[
Y=s(X)+r(X)\operatorname{expit}(b^\top T)+\varepsilon,
\qquad
\varepsilon\sim \mathrm{N}(0,\sigma_{\varepsilon}^2),\quad
\varepsilon\perp(X,T).
\]
Here \(\sigma_{\varepsilon}=0.1\).
Since \(\mathbb{E}\{\operatorname{expit}(b^\top T)\}=1/2\), the target regression function is
\[
f^*(x)=s(x)+\frac12 r(x).
\]
The following two constructions specify different levels of complexity for \(s(\cdot)\) and \(r(\cdot)\). Let \(\mathcal J=(j_1,\ldots,j_8)\) be a coordinate tuple. Define
\begin{align*}
	g_{\mathrm{e}}(x;\mathcal J)
	=&
	\sin(x_{j_1})+0.5x_{j_2}+0.25x_{j_3}^2,\quad \text{and}\\
	g_{\mathrm{h}}(x;\mathcal J)
	=&
	\sin(x_{j_1}x_{j_2})
	+0.5\cos(x_{j_3}+x_{j_4})
	+0.5x_{j_5}x_{j_6}
	+0.25|x_{j_7}|
	+0.25x_{j_8}^2.
\end{align*}
These functions are centered and rescaled using Monte Carlo estimates; the same notation denotes their standardized versions. Set
$\mathcal J_s=(1,\ldots,8)$,
$\mathcal J_r=(9,\ldots,16)$, and
$\mathcal J_q=(17,\ldots,24)$.
We consider four data-generating processes (DGPs), corresponding to the combinations of
$s(x)\in
\{g_{\mathrm{e}}(x;\mathcal J_s),g_{\mathrm{h}}(x;\mathcal J_s)\}$ and $r(x)\in
\{g_{\mathrm{e}}(x;\mathcal J_r),g_{\mathrm{h}}(x;\mathcal J_r)\}$, denoted by EE, EH, HE, and HH.
The first and second letters indicate the choices for \(s(\cdot)\) and \(r(\cdot)\), respectively, with E and H corresponding to \(g_{\mathrm{e}}\) and \(g_{\mathrm{h}}\).

\smallskip
\noindent \textbf{Pseudo-label mechanisms.} Table~\ref{tab:simulation-pseudo-label-mechanisms} lists the seven pseudo-label mechanisms and their corresponding calibration functions \(\mu^*\).
\begin{table}[htbp]
\centering
\caption{Pseudo-label mechanisms.}
\label{tab:simulation-pseudo-label-mechanisms}
\footnotesize
\setlength{\tabcolsep}{4pt}
\renewcommand{\arraystretch}{1.10}
\begin{tabular}{@{}lll@{}}
\toprule
\textbf{Mechanism} & {\bfseries\boldmath \(\widetilde Y\)} & {\bfseries\boldmath \(\mu^*(x,\tilde y)\)} \\
\midrule
\textbf{NORS} --- Noiseless response signal & \(s(X)+r(X)\operatorname{expit}(b^\top T)\) & \(\tilde y\) \\
\midrule[0.3pt]
\textbf{AURS} --- Auxiliary-response signal & \(r(X)\operatorname{expit}(b^\top T)\) & \(s(x)+\tilde y\) \\
\textbf{RARS} --- Reversed auxiliary-response signal & \(-r(X)\operatorname{expit}(b^\top T)\) & \(s(x)-\tilde y\) \\
\textbf{CARS} --- Contaminated auxiliary-response signal & \(r(X)\operatorname{expit}(b^\top T)+q(X)\) & \(s(x)+\tilde y-q(x)\) \\
\midrule[0.3pt]
\textbf{RAAS} --- Raw auxiliary signal & \(\operatorname{expit}(b^\top T)\) & \(s(x)+\tilde y r(x)\) \\
\textbf{COAS} --- Coarsened auxiliary signal & \(\mathbf 1\{\operatorname{expit}(b^\top T)>1/2\}\) & \(s(x)+c_{\tilde y}r(x)\) \\
\midrule[0.3pt]
\textbf{INDN} --- Independent noise & \(W,\ W\sim\mathrm{N}(0,1)\) & \(s(x)+(1/2)r(x)\) \\
\bottomrule
\end{tabular}
\par\smallskip
{\footnotesize\raggedright
\(W\perp(X,T,Y)\), \(q(x)=g_{\mathrm{e}}(x;\mathcal J_q)\), and
\(c_a=\mathbb E\{\operatorname{expit}(b^\top T)\mid
\mathbf 1\{\operatorname{expit}(b^\top T)>1/2\}=a\}\), \(a\in\{0,1\}\).
\par}
\end{table}

\subsubsection{Overall Performance}
Table~\ref{tab:simulation-overall-performance} reports the mean relative improvements in test excess risk over the supervised estimator, \(\mathit{RI}_m\), across DGPs and pseudo-label mechanisms.
The two fixed-weight wPPRC estimators and VS show broadly similar performance and achieve the largest mean improvements in most settings. Both wPPRC(\(N_\ell/N\)) and VS achieve positive mean improvements in all 28 settings, while wPPRC(0) does so in 27 settings. Although PL performs best in a few settings, PL and PPI yield negative mean improvements in 20 and 24 settings, respectively.
Results for \(\widehat R_m\) and \(\mathit{RIR}_m\) are reported in Section~S7.1.3 of the SM.

\begin{table}[!htbp]
\color{black}
\centering
\caption{Relative improvements (\%), reported as mean (SE). Bold and underlined values indicate the best and second-best means, respectively.}
\label{tab:simulation-overall-performance}
\footnotesize
\setlength{\tabcolsep}{2pt}
\renewcommand{\arraystretch}{1.15}
\resizebox{\linewidth}{!}{%
\begin{tabular}{@{}l*{5}{c}@{\hspace{12pt}}*{5}{c}@{}}
\toprule
& \multicolumn{5}{c}{\textbf{DGP: EE}} & \multicolumn{5}{c}{\textbf{DGP: EH}} \\
& \multicolumn{5}{c}{\(\widehat{\mathcal E}_{\mathrm{Sup}}=0.326\,(0.002)\)} & \multicolumn{5}{c}{\(\widehat{\mathcal E}_{\mathrm{Sup}}=0.557\,(0.003)\)} \\
\cmidrule(lr){2-6}\cmidrule(l){7-11}
\textbf{Pseudo-label} & \textbf{wPPRC(0)} & \textbf{wPPRC(\(N_\ell/N\))} & \textbf{VS} & \textbf{PL} & \textbf{PPI} & \textbf{wPPRC(0)} & \textbf{wPPRC(\(N_\ell/N\))} & \textbf{VS} & \textbf{PL} & \textbf{PPI} \\
\midrule
NORS & \tblval{74.26}{0.25} & \tblval{74.35}{0.24} & \tblval{\underline{75.14}}{0.23} & \tblval{\textbf{78.88}}{0.20} & \tblval{58.23}{0.50} & \tblval{72.79}{0.23} & \tblval{73.08}{0.24} & \tblval{\underline{73.55}}{0.22} & \tblval{\textbf{76.82}}{0.23} & \tblval{60.42}{0.34} \\
AURS & \tblval{\underline{18.05}}{0.57} & \tblval{17.45}{0.55} & \tblval{\textbf{18.22}}{0.61} & \tblval{-193.22}{1.92} & \tblval{-126.99}{2.91} & \tblval{41.76}{0.42} & \tblval{\underline{42.40}}{0.41} & \tblval{\textbf{42.57}}{0.48} & \tblval{-81.11}{1.19} & \tblval{-63.80}{1.91} \\
RARS & \tblval{18.07}{0.57} & \tblval{\underline{18.22}}{0.54} & \tblval{\textbf{18.37}}{0.55} & \tblval{-356.11}{3.10} & \tblval{-204.28}{3.58} & \tblval{41.80}{0.40} & \tblval{\textbf{42.35}}{0.40} & \tblval{\underline{42.19}}{0.43} & \tblval{-124.90}{1.45} & \tblval{-91.41}{3.52} \\
CARS & \tblval{4.44}{0.64} & \tblval{\underline{4.59}}{0.59} & \tblval{\textbf{5.97}}{0.62} & \tblval{-228.01}{2.55} & \tblval{-167.16}{5.48} & \tblval{19.00}{0.47} & \tblval{\underline{19.55}}{0.45} & \tblval{\textbf{19.72}}{0.47} & \tblval{-110.41}{1.96} & \tblval{-79.29}{1.96} \\
RAAS & \tblval{1.37}{0.64} & \tblval{\underline{1.51}}{0.56} & \tblval{\textbf{3.39}}{0.51} & \tblval{-356.41}{3.06} & \tblval{-153.30}{5.11} & \tblval{\underline{2.34}}{0.47} & \tblval{1.94}{0.46} & \tblval{\textbf{2.40}}{0.51} & \tblval{-146.38}{1.63} & \tblval{-69.34}{3.06} \\
COAS & \tblval{2.37}{0.58} & \tblval{\underline{2.41}}{0.54} & \tblval{\textbf{3.16}}{0.57} & \tblval{-360.08}{3.52} & \tblval{-169.21}{7.31} & \tblval{\textbf{2.40}}{0.48} & \tblval{1.64}{0.46} & \tblval{\underline{1.72}}{0.53} & \tblval{-148.84}{2.29} & \tblval{-74.79}{2.35} \\
INDN & \tblval{-0.15}{0.64} & \tblval{\underline{0.45}}{0.58} & \tblval{\textbf{2.47}}{0.50} & \tblval{-313.99}{3.05} & \tblval{-200.36}{3.66} & \tblval{\textbf{2.66}}{0.48} & \tblval{\underline{2.45}}{0.50} & \tblval{1.98}{0.51} & \tblval{-122.18}{1.49} & \tblval{-87.78}{1.64} \\
\midrule
& \multicolumn{5}{c}{\textbf{DGP: HE}} & \multicolumn{5}{c}{\textbf{DGP: HH}} \\
& \multicolumn{5}{c}{\(\widehat{\mathcal E}_{\mathrm{Sup}}=1.167\,(0.006)\)} & \multicolumn{5}{c}{\(\widehat{\mathcal E}_{\mathrm{Sup}}=1.351\,(0.005)\)} \\
\cmidrule(lr){2-6}\cmidrule(l){7-11}
\textbf{Pseudo-label} & \textbf{wPPRC(0)} & \textbf{wPPRC(\(N_\ell/N\))} & \textbf{VS} & \textbf{PL} & \textbf{PPI} & \textbf{wPPRC(0)} & \textbf{wPPRC(\(N_\ell/N\))} & \textbf{VS} & \textbf{PL} & \textbf{PPI} \\
\midrule
NORS & \tblval{83.54}{0.14} & \tblval{83.72}{0.14} & \tblval{\textbf{84.12}}{0.15} & \tblval{\underline{83.98}}{0.21} & \tblval{71.26}{0.25} & \tblval{78.87}{0.17} & \tblval{\underline{79.57}}{0.14} & \tblval{\textbf{79.73}}{0.13} & \tblval{72.28}{0.39} & \tblval{66.52}{0.27} \\
AURS & \tblval{\underline{9.01}}{0.49} & \tblval{8.97}{0.49} & \tblval{9.00}{0.49} & \tblval{\textbf{11.84}}{0.44} & \tblval{-5.10}{0.81} & \tblval{15.41}{0.40} & \tblval{15.37}{0.40} & \tblval{\underline{15.85}}{0.44} & \tblval{\textbf{18.79}}{0.32} & \tblval{-34.28}{34.73} \\
RARS & \tblval{9.31}{0.52} & \tblval{\underline{9.48}}{0.50} & \tblval{\textbf{9.70}}{0.54} & \tblval{-14.00}{0.61} & \tblval{-19.54}{7.25} & \tblval{\underline{15.45}}{0.47} & \tblval{15.28}{0.46} & \tblval{\textbf{15.83}}{0.44} & \tblval{4.66}{0.38} & \tblval{-12640.86}{12429.67} \\
CARS & \tblval{6.06}{0.52} & \tblval{\underline{6.08}}{0.51} & \tblval{\textbf{6.91}}{0.49} & \tblval{-14.51}{0.81} & \tblval{-9.28}{1.74} & \tblval{8.31}{0.42} & \tblval{\textbf{8.59}}{0.40} & \tblval{\underline{8.55}}{0.39} & \tblval{-1.70}{0.53} & \tblval{-2.42}{3.96} \\
RAAS & \tblval{\underline{5.28}}{0.50} & \tblval{4.97}{0.47} & \tblval{\textbf{6.49}}{0.50} & \tblval{-18.79}{0.75} & \tblval{-14.13}{4.24} & \tblval{4.72}{0.35} & \tblval{\underline{5.13}}{0.35} & \tblval{\textbf{6.22}}{0.40} & \tblval{-3.91}{0.54} & \tblval{-7860.91}{7593.58} \\
COAS & \tblval{5.38}{0.52} & \tblval{\underline{5.41}}{0.50} & \tblval{\textbf{6.45}}{0.50} & \tblval{-18.78}{0.84} & \tblval{-13.46}{5.41} & \tblval{4.71}{0.39} & \tblval{\underline{5.12}}{0.40} & \tblval{\textbf{6.08}}{0.40} & \tblval{-4.07}{0.68} & \tblval{-25.41}{15.13} \\
INDN & \tblval{5.42}{0.50} & \tblval{\underline{5.50}}{0.49} & \tblval{\textbf{6.76}}{0.48} & \tblval{-9.76}{0.80} & \tblval{-184.99}{173.86} & \tblval{4.97}{0.33} & \tblval{\underline{5.25}}{0.34} & \tblval{\textbf{6.64}}{0.35} & \tblval{3.65}{0.60} & \tblval{-55.69}{46.06} \\
\bottomrule
\end{tabular}%
}
\end{table}

Under NORS, the pseudo-label provides an ideal prediction of the response, and all methods attain their largest mean relative improvements within each DGP. PL achieves the largest mean improvement in EE and EH, while VS achieves the largest in HE and HH. The two fixed-weight wPPRC estimators remain close to the best-performing method. These results show that wPPRC and VS remain competitive even in settings where highly accurate pseudo-labels favor PL. PPI also achieves substantial mean improvements over the supervised estimator, ranging from 58.2\% to 71.3\% across the four DGPs.

Under AURS, RARS, and CARS, pseudo-labels contain the \(r(X)\)-related component, eliminating the need to learn \(r(X)\) in the first step. AURS and RARS differ only in sign and contain equivalent response information conditional on \(X\).
Reversing the pseudo-label sign from AURS to RARS has little effect on the mean relative improvements of wPPRC and VS but reduces those of PL.
Under AURS, PL achieves the largest mean relative improvement in HE and HH, but this advantage disappears under RARS.
The mean relative improvements of wPPRC and VS are smaller with CARS than with AURS in every DGP. This may reflect the additional complexity of calibration introduced by the contaminating component in CARS.
Across these mechanisms, wPPRC and VS benefit from pseudo-labels that retain useful response information, even when they are sign-reversed or contaminated.

RAAS and COAS provide auxiliary information, but calibration still requires learning both \(s(X)\) and \(r(X)\) from labeled data. This may explain why wPPRC and VS achieve smaller mean relative improvements than under AURS and RARS in every DGP.
Under INDN, the pseudo-label provides no additional response information conditional on \(X\), yet wPPRC and VS generally retain modest positive mean improvements, except for wPPRC(0) in EE, whose mean relative improvement is \(-0.15\%\).

To examine how the difficulty of estimating each response component affects the gains of wPPRC and VS, we compare DGPs under AURS and RARS. Holding \(s(\cdot)\) fixed, changing \(r(\cdot)\) from easy to hard increases their mean relative improvements, as seen in the EE--EH and HE--HH comparisons. Under AURS and RARS, the pseudo-label already contains the \(r(X)\)-related component, making it more useful when this component is harder to estimate from labeled data alone. Holding \(r(\cdot)\) fixed, changing \(s(\cdot)\) from easy to hard decreases their mean relative improvements, as seen in the EE--HE and EH--HH comparisons. Since \(s(X)\) must be learned from labeled data, greater estimation difficulty may limit the gains. These comparisons suggest that wPPRC and VS gain more when pseudo-labels capture the harder-to-estimate response component, consistent with Section~\ref{sec:relu-upper-bound}.

For the neural-network PPI implementation considered here, the mean relative improvements are positive under NORS but negative under all other mechanisms. Large negative mean relative improvements are accompanied by large standard errors in HH–RARS and HH–RAAS. In HH–RARS, one extreme run drives the mean relative improvement of \(-12{,}640.86\%\). The potential instability of the rectified neural-network objective is discussed in Section~\ref{sec:connections-existing-methods} and Section~S6.3 of the SM.

\subsubsection{Weight Selection}
\begin{figure}[!htbp]
\centering
\includegraphics[width=\textwidth]{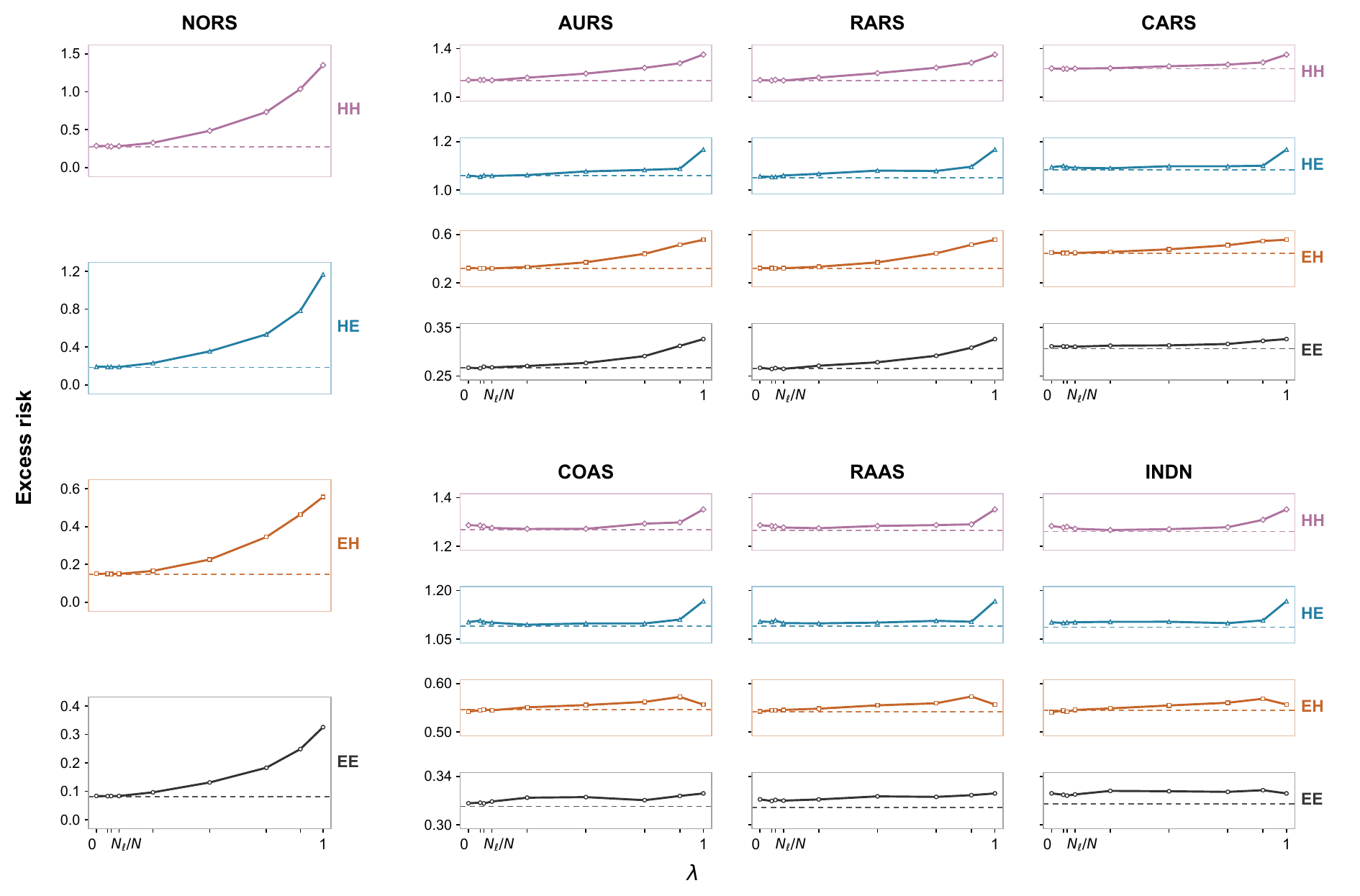}
\caption{Mean test excess risk versus $\lambda$. Dashed lines show the mean test excess risk of the VS estimator.}
\label{fig:weight-selection-excess-risk}
\end{figure}

Figure~\ref{fig:weight-selection-excess-risk} shows mean test excess risk as a function of \(\lambda\); dashed lines indicate the corresponding values for VS. Under NORS, mean test excess risk is low and changes little at small weights, then increases as \(\lambda\) approaches 1. This pattern is consistent with the benefit of using abundant unlabeled data when calibration is simple. A similar pattern appears under AURS and RARS, particularly in EH and HH, while the trend is generally weaker under CARS. For COAS, RAAS, and INDN, mean test excess risk varies less across weights.

Table~\ref{tab:weight-selection-counts} reports how often each weight is selected over 100 replications. Under NORS, selection concentrates on small weights, with \(\lambda\le0.1\) selected in 92--99 replications in each DGP. Under AURS and RARS, selection favors small weights in EH and HH but is more dispersed in EE and HE, where intermediate weights are frequently selected. Under CARS, selection favors small weights in EH but shifts toward intermediate or larger weights in the other DGPs. For RAAS, COAS, and INDN, weights \(\lambda\ge0.5\) are selected in roughly half or more of the replications in every DGP.

\begin{table}[!htbp]
\centering
\caption{Weight-selection counts over 100 replications. Within each setting, the highest counts are bold, including ties.}
\label{tab:weight-selection-counts}
\footnotesize
\setlength{\tabcolsep}{2pt}
\renewcommand{\arraystretch}{1.15}
\resizebox{0.8\linewidth}{!}{%
\begin{tabular}{@{}l*{9}{c}@{\hspace{12pt}}*{9}{c}@{}}
\toprule
 & \multicolumn{9}{c}{\textbf{DGP: EE}} & \multicolumn{9}{c}{\textbf{DGP: EH}} \\
\cmidrule(l{2pt}r{12pt}){2-10}\cmidrule{11-19}
\textbf{Pseudo-label} & $0$ & $0.05$ & $N_\ell/N$ & $0.1$ & $0.25$ & $0.5$ & $0.75$ & $0.9$ & $1$ & $0$ & $0.05$ & $N_\ell/N$ & $0.1$ & $0.25$ & $0.5$ & $0.75$ & $0.9$ & $1$ \\
\midrule
NORS & 24 & \textbf{28} & 23 & 24 & 1 & 0 & 0 & 0 & 0 & 21 & 19 & 23 & \textbf{29} & 8 & 0 & 0 & 0 & 0 \\
AURS & \textbf{21} & 11 & 12 & 14 & 15 & 18 & 5 & 2 & 2 & 13 & \textbf{27} & 22 & 22 & 13 & 3 & 0 & 0 & 0 \\
RARS & 14 & \textbf{21} & 13 & 16 & 16 & 9 & 6 & 5 & 0 & 19 & 22 & 19 & \textbf{27} & 11 & 2 & 0 & 0 & 0 \\
CARS & 14 & 7 & 9 & 7 & 12 & 15 & \textbf{17} & 11 & 8 & \textbf{23} & 21 & 21 & 14 & 15 & 5 & 1 & 0 & 0 \\
COAS & 11 & 7 & 6 & 14 & 8 & 8 & 16 & \textbf{21} & 9 & 11 & 14 & 7 & 10 & 7 & 11 & 13 & \textbf{15} & 12 \\
RAAS & 10 & 7 & 7 & 8 & 13 & 9 & 10 & \textbf{23} & 13 & 10 & 13 & 5 & 12 & 10 & 7 & 13 & 8 & \textbf{22} \\
INDN & 8 & 8 & 7 & 8 & 7 & 10 & 18 & \textbf{20} & 14 & 10 & 10 & 9 & 11 & 11 & 10 & 9 & \textbf{15} & \textbf{15} \\
\midrule
 & \multicolumn{9}{c}{\textbf{DGP: HE}} & \multicolumn{9}{c}{\textbf{DGP: HH}} \\
\cmidrule(l{2pt}r{12pt}){2-10}\cmidrule{11-19}
\textbf{Pseudo-label} & $0$ & $0.05$ & $N_\ell/N$ & $0.1$ & $0.25$ & $0.5$ & $0.75$ & $0.9$ & $1$ & $0$ & $0.05$ & $N_\ell/N$ & $0.1$ & $0.25$ & $0.5$ & $0.75$ & $0.9$ & $1$ \\
\midrule
NORS & 23 & 26 & \textbf{28} & 22 & 1 & 0 & 0 & 0 & 0 & 25 & 22 & \textbf{27} & 24 & 2 & 0 & 0 & 0 & 0 \\
AURS & \textbf{17} & 9 & 9 & 11 & 10 & 9 & 15 & 13 & 7 & 20 & 20 & 18 & \textbf{25} & 12 & 4 & 0 & 0 & 1 \\
RARS & 16 & \textbf{17} & 5 & 13 & 9 & \textbf{17} & 8 & 11 & 4 & \textbf{25} & 17 & 17 & 20 & 12 & 3 & 4 & 2 & 0 \\
CARS & 8 & 5 & 12 & 8 & 8 & 10 & 13 & \textbf{27} & 9 & 13 & 11 & 10 & 11 & 14 & 11 & \textbf{16} & 12 & 2 \\
COAS & 9 & 7 & 6 & 8 & 11 & 16 & 12 & \textbf{20} & 11 & 7 & 6 & 3 & 8 & 14 & \textbf{23} & 12 & 17 & 10 \\
RAAS & 8 & 7 & 7 & 6 & 16 & 13 & \textbf{18} & \textbf{18} & 7 & 7 & 7 & 6 & 13 & 13 & 13 & \textbf{19} & 16 & 6 \\
INDN & 11 & 4 & 7 & 12 & 8 & 18 & 11 & \textbf{20} & 9 & 3 & 1 & 7 & 7 & 18 & \textbf{24} & 22 & 15 & 3 \\
\bottomrule
\end{tabular}
}
\end{table}

\subsection{Predicting Homeless Counts}
\label{sec:real-data-analysis}

We consider the prediction of homeless counts across census tracts in Los Angeles County.
Local estimates of homeless populations help agencies assess service needs and allocate resources \citep{hud2014pit}.
However, obtaining these counts requires field visits, and the cost of surveying a large metropolitan area limits the number of tracts that can be visited.
The 2004--2005 Los Angeles County homeless survey recorded homeless counts in selected census tracts, with covariate information available for all tracts \citep{berk2008counting,kriegler2010small}.\footnote{These data were also analyzed by \citet{zhang2019semi} in the context of semi-supervised mean estimation. We thank Professor Anru Zhang for kindly providing the homeless survey data.}
Recent advances in tabular foundation models offer a new opportunity to address this prediction problem \citep{vanbreugel2024priority,hollmann2025accurate}.
We investigate whether our method improves tract-level prediction with limited labeled data using pseudo-labels from these models.

The response \(Y\) is the homeless count in each census tract.
The covariate vector \(X\) comprises seven variables: \texttt{Perc.\allowbreak Industrial}, \texttt{Perc.\allowbreak Residential}, \texttt{Perc.\allowbreak Vacant}, \texttt{Perc.\allowbreak Commercial}, \texttt{Perc.\allowbreak OwnerOcc}, \texttt{Perc.\allowbreak Minority}, and \texttt{Median\allowbreak Household\allowbreak Income}.
We use 265 randomly selected surveyed tracts and \(N_u=1{,}545\) unsurveyed tracts.
The surveyed tracts are randomly divided into training, tuning, validation, and test sets of sizes \(N_\ell=148\), \(N_{\ell,\mathrm{tune}}=32\), \(N_{\ell,\mathrm{val}}=32\), and \(N_{\mathrm{test}}=53\), respectively.

Pseudo-labels are generated through in-context learning using 24 tabular foundation model configurations, including TabPFN \citep{jaeger2026tabpfn35}, TabICL \citep{qu2026tabiclv2}, Mitra \citep{zhang2025mitra}, Nori \citep{synthefy2026nori}, LimiX \citep{zhang2026limix2}, Xiaomi-TabLDM \citep{wang2026xiaomitabldm} and their variants.
For each model, pseudo-labels for labeled training observations are generated using the remaining training observations as context (leave-one-out), whereas pseudo-labels for unlabeled observations use the full labeled training sample as context. Implementation details are provided in the SM.

\begin{figure}[!htbp]
\centering
\includegraphics[width=\textwidth]{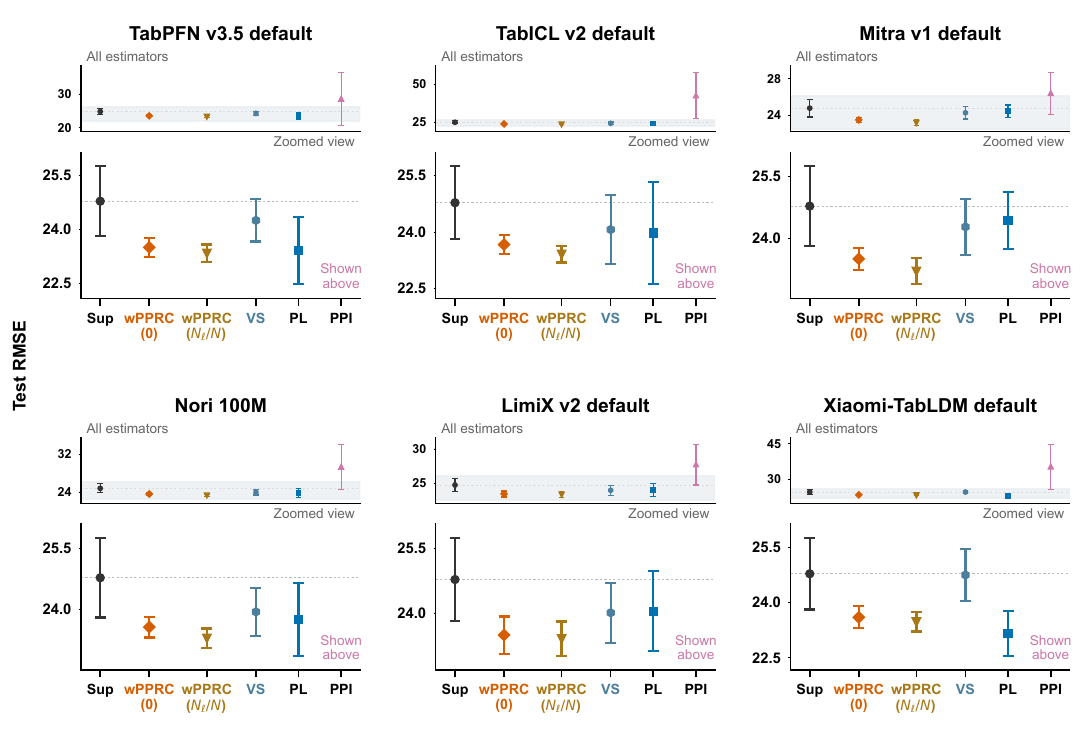}
\caption{Test RMSE on the Homeless dataset for six representative configurations. Points and error bars show means \(\pm1.96\) SE over 20 training seeds. Lower panels enlarge shaded regions; dashed lines indicate Sup. Panel scales differ.}
\label{fig:homeless-rmse}
\end{figure}

We compare the same six estimators as in Section~\ref{sec:simulation-studies}, using the same candidate weight grid for VS.
Performance is evaluated by test root mean squared error (RMSE) on the original count scale for each training seed.
Figure~\ref{fig:homeless-rmse} compares six representative configurations, one from each model family; results for all 24 configurations appear in the SM.

Across all 24 configurations, wPPRC(0) and wPPRC(\(N_\ell/N\)) show similar performance and both achieve lower mean test RMSE than Sup, with mean seedwise relative RMSE reductions ranging from 3.8\% to 6.4\% and from 4.0\% to 6.0\%, respectively. Both estimators also have lower mean RMSE than PPI in all 24 configurations. Against PL, wPPRC(0) yields lower mean RMSE in 16 configurations and wPPRC(\(N_\ell/N\)) in 19. For example, with pseudo-labels from Mitra v1 (default), the mean test RMSE is 23.497 for wPPRC(0) and 23.196 for wPPRC(\(N_\ell/N\)), compared with 24.433 for PL and 26.402 for PPI. VS achieves lower mean test RMSE than Sup in 20 configurations and than PPI in all 24, but performs less favorably than the two fixed-weight wPPRC estimators and PL. This may reflect the difficulty of selecting among nine candidate weights using only 32 validation observations.

Both fixed-weight wPPRC estimators show relatively stable performance across training seeds. Their standard errors of test RMSE range from 0.124 to 0.262 for wPPRC(0) and from 0.103 to 0.318 for wPPRC(\(N_\ell/N\)), smaller than those of the other estimators in most configurations. VS exhibits greater variability than either fixed-weight estimator, possibly due to the limited validation sample, as discussed above. PL’s standard errors vary more widely across configurations, ranging from 0.215 to 1.137, whereas PPI has the largest values across all 24 configurations, ranging from 1.165 to 12.827.

Overall, these results illustrate the practical value of the wPPRC framework for using pseudo-labels to improve tract-level prediction with limited surveyed data.

\section{Conclusion}
\label{sec:conclusion}
The value of pseudo-labels depends not only on their predictive accuracy but also on how readily their relationship with the response can be learned. Calibration can exploit this relationship even when pseudo-labels are biased. To quantify the value of additional pseudo-labels, we compare minimax risks across observation structures. Observing pseudo-labels alongside additional covariates can yield gains unavailable under the baseline structures, and these gains can persist after validation-based weight selection even when the baselines receive the same total labeled-data budget.

Several directions merit further study: (1) In our real-data application, tabular foundation models generate pseudo-labels using leave-one-out contexts for labeled observations and the full training context for unlabeled observations. The overlap among these contexts induces dependencies that are not directly covered by our current i.i.d. framework. This motivates extending the theory to data-dependent pseudo-label generation.
(2) Different models may provide complementary information, motivating methods for selecting or combining pseudo-labels from multiple sources.
(3) Differences between the source and target populations may affect the usefulness of pseudo-labels, raising questions about when their benefits persist and how calibration should adapt.
(4) The analysis could also be extended to temporal, spatial, and within-cluster dependence to clarify how these dependencies affect the gains from additional observations.

\paragraph*{Acknowledgements.}
This work is supported by the Research Grants Council (RGC) of Hong Kong
(P0053801), the Joint Laboratory of Applied Mathematics (P0052683), and
a research grant from The Hong Kong Polytechnic University (P0046811).
The Joint Laboratory is funded through the RGC's Co-funding Mechanism on
Joint Laboratories with the Chinese Academy of Sciences (CAS).

\bibliographystyle{apalike-six}
\bibliography{pprc}

\end{document}